\documentclass{article}

\usepackage{bbm}
\usepackage{caption}
\usepackage{rotating}
\usepackage[maxfloats=100]{morefloats}
\usepackage{listings}
\usepackage{booktabs}
\usepackage{xcolor}
\usepackage[title]{appendix}
\usepackage{makecell} 
\usepackage{multirow}

\usepackage[symbol]{footmisc}

\usepackage{longtable}
\usepackage{subcaption}
\usepackage[numbers]{natbib}
\usepackage{geometry}
\usepackage{amssymb}
\usepackage{amsthm}
\usepackage{enumitem}
\usepackage{verbatim}
\usepackage{xcolor}
\usepackage{graphicx}
\usepackage{enumitem}
\usepackage{comment}
\usepackage{subfiles}
\usepackage{amsmath}
\usepackage{todonotes}
\usepackage{tikz-qtree,tikz-qtree-compat}
\usepackage{scrextend}
\usepackage{framed}
\usepackage{placeins}
\usepackage{setspace}
\usepackage{mathtools,collcell,eqparbox}
\usepackage{ dsfont }
\usepackage{multicol}
\usepackage{pgfplots}
\usepackage{tikz}
\usepackage{listings}
\usepackage{courier}
\usepackage{url}
\usepackage{color}
\usepackage[title]{appendix}
\usepackage{authblk}
\usepackage[pagebackref = true]{hyperref}
\hypersetup{
    colorlinks = true,
    linkcolor = teal,
    anchorcolor = teal,
    citecolor = teal,
    filecolor = teal,
    urlcolor = teal
    }

\pgfplotsset{compat=1.15}
\usepackage{algorithm}
\usepackage{algpseudocode}
\usepackage{amsmath}
\usepackage{amssymb}
\usepackage{amsthm}
\usepackage{graphicx}
\usepackage{multicol}
\usepackage[title]{appendix}
\usepackage{makecell} 
\usepackage{multirow}
\usepackage{todonotes}
\usepackage{ dsfont }
\usepackage{tablefootnote}

\newtheorem{theorem}{Theorem}
\newtheorem{assumption}{Assumption}

\newtheorem{proposition}{Proposition}[section]

\newtheorem{lemma}{Lemma}
\theoremstyle{definition}
\newtheorem{example}{Example}
\newtheorem{definition}{Definition}
\theoremstyle{remark}
\newtheorem*{remark}{Remark}

\newcommand{\inv}{^{-1}}

\newcommand{\1}{\mathds{1}}

\newcommand{\bd}[1]{\mathbf{#1}}
\newcommand{\mrm}[1]{\mathrm{#1}}

\newcommand{\diam}{\mathrm{diam}}

\newcommand{\PiH}{\Pi_{\mathrm{H}}}
\newcommand{\PiM}{\Pi_{\mathrm{M}}}
\newcommand{\PiS}{\Pi_{\mathrm{S}}}

\newcommand{\GammaH}{\Gamma_{\mathrm{H}}}
\newcommand{\GammaS}{\Gamma_{\mathrm{S}}}
\newcommand{\GammaM}{\Gamma_{\mathrm{M}}}

\newcommand{\barGammaH}{\overline\Gamma_{\mathrm{H}}}
\newcommand{\barGammaS}{\overline\Gamma_{\mathrm{S}}}
\newcommand{\barGammaM}{\overline\Gamma_{\mathrm{M}}}

\newcommand{\A}{\mathbb{A}}
\newcommand{\W}{\mathbb{W}}
\newcommand{\R}{\mathbb{R}}

\newcommand{\X}{\mathbb{X}}
\newcommand{\C}{\mathbb{C}}

\newcommand{\Z}{\mathbb{Z}}

\newcommand{\ra}{\rightarrow}
\newcommand{\Ra}{\Rightarrow}

\newcommand{\cd}{\cdot}
\newcommand{\ds}{\cdots}

\newcommand{\cA}{\mathcal{A}}
\newcommand{\cB}{\mathcal{B}}
\newcommand{\cC}{\mathcal{C}}
\newcommand{\cD}{\mathcal{D}}

\newcommand{\cF}{\mathcal{F}}
\newcommand{\cG}{\mathcal{G}}

\newcommand{\cK}{\mathcal{K}}

\newcommand{\cN}{\mathcal{N}}

\newcommand{\cP}{\mathcal{P}}
\newcommand{\cQ}{\mathcal{Q}}
\newcommand{\cR}{\mathcal{R}}

\newcommand{\cT}{\mathcal{T}}
\newcommand{\cU}{\mathcal{U}}

\newcommand{\cW}{\mathcal{W}}
\newcommand{\cX}{\mathcal{X}}

\newcommand{\TV}[1]{\left\|#1\right\|_\mathrm{TV}}

\newcommand{\set}[1]{\left\{{#1}\right\}}

\newcommand{\ceil}[1]{\left\lceil{#1}\right\rceil}

\newcommand{\norm}[1]{\left\|#1\right\|}

\newcommand{\abs}[1]{\left|#1\right|}
\newcommand{\sqbk}[1]{\left[ #1 \right]}
\newcommand{\sqbkcond}[2]{\left[ #1 \middle| #2 \right]}

\newcommand{\crbk}[1]{\left( #1 \right)}

\numberwithin{equation}{section}

\title{Learning Optimal Distributionally Robust Stochastic Control \\
in Continuous State Spaces}

\author[1]{Shengbo Wang}
\author[2]{Jason Meng}
\author[3]{Nian Si}
\author[2]{Jose Blanchet}
\author[4]{Zhengyuan Zhou}

\affil[1]{Daniel J. Epstein Department of Industrial and Systems Engineering\\ USC}

\affil[2]{Management Science and Engineering\\Stanford University}
\affil[3]{Department of Industrial Engineering and Decision Analytics\\
HKUST}
\affil[4]{Stern School of Business, New York University}
\date{November 2025}

\begin{document}

\maketitle

\begin{abstract}
We study data-driven learning of robust stochastic control for infinite-horizon systems with potentially continuous state and action spaces. In many managerial settings--supply chains, finance, manufacturing, services, and dynamic games--the state-transition mechanism is determined by system design, while available data capture the distributional properties of the stochastic inputs from the environment. For modeling and computational tractability, a decision maker often adopts a Markov control model with i.i.d. environment inputs, which can render learned policies fragile to internal dependence or external perturbations. We introduce a distributionally robust stochastic control paradigm that promotes policy reliability by introducing adaptive adversarial perturbations to the environment input, while preserving the modeling, statistical, and computational tractability of the Markovian formulation. From a modeling perspective, we examine two adversary models--current-action-aware and current-action-unaware--leading to distinct dynamic behaviors and robust optimal policies. From a statistical learning perspective, we characterize optimal finite-sample minimax rates for uniform learning of the robust value function across a continuum of states under ambiguity sets defined by the $f_k$-divergence and Wasserstein distance. To efficiently compute the optimal robust policies, we further propose algorithms inspired by deep reinforcement learning methodologies. Finally, we demonstrate the applicability of the framework to real managerial problems.
\end{abstract}

\section{Introduction}\label{sec:Intro}

Stochastic optimal control formulations are extensively utilized in the modeling, design, and optimization of systems influenced by probabilistic dynamics. These formulations play a crucial role across various fields within operations research and management disciplines. Notable applications of stochastic control can be seen in finance \citep{merton1976pricing_with_jump}, supply chain management \citep{wagner2018robust}, communication systems \citep{yuksel2013networked_control}, manufacturing and operations management \citep{tse1997admission_control, porteus2002inventory}, as well as energy systems \citep{foschini1993power_control}. Modeling such systems naturally leads to the use of continuous state spaces, supporting high-resolution dynamic models that accurately capture the underlying system dynamics. 

The state transition dynamics of a broad class of stochastic control problems can be expressed through the following recursive relation
\begin{equation}\label{eqn:state_dynamics}
    X_{t+1} = f(X_t, A_t, W_t).
\end{equation} In this expression, \( X_t,A_t \) are the state of the system and the action of the controller at time \( t \), respectively, and \( \{W_t : t \geq 0\} \) is a sequence of random variables, usually representing the stochastic inputs from the environment, that underlie the stochasticity of the system.

In infinite-horizon formulations, classical Markov stochastic control and MDP models assume that the environment inputs $\{W_t\}$ form an independent and identically distributed sequence \citep{bertsekas2012DP}. This assumption is central to classical dynamic programming theories because it ensures the optimality of stationary Markov policies. Under it, the state–action process $\{(X_t,A_t):t\geq 0\}$ is referred to as a controlled Markov chain, and the Bellman equation characterizes both the optimal control value and the corresponding stationary policy. These properties enable both value- and policy-based approximate dynamic programming and reinforcement learning methods—such as value/Q-iteration with function approximation, policy-gradient, and actor–critic approaches—to be applied effectively even in environments with continuous state and action spaces.

The i.i.d. premise is, however, often violated in managerial environments. In practice, demand, arrivals, and yields exhibit temporal correlations, seasonality, delayed effects, or regime shifts \citep{rohde2024intertemporal}; competitive interactions and pricing introduce strategic responses in market dynamics \citep{kim2018competitive}; and endogeneity, confounding, and history-dependence are common in feedback-driven systems \citep{kallus2021minimax}. When transition dynamics are modeled or learned under an i.i.d. premise, they are typically misspecified, and the resulting policies may become unreliable in deployment, particularly in infinite-horizon settings where the very optimality of a policy can induce fragility in dynamic decision-making environments \citep{fan2024fragility}. Small systematic errors in the model can accumulate over time, leading to inflated inventories, longer queues and waiting times, and, in some cases, unstable controls.

We therefore develop a distributionally robust stochastic control (DRSC) framework that directly mitigates the risks arising from a misspecified i.i.d. assumption, while preserving the learning and deployment tractability of the classical model. To induce robustness, the framework adopts an adversarial perspective \citep{pinto2017robust}, requiring the decision maker to plan and control optimally in the presence of an adversary capable of perturbing the sequence of environmental inputs. This leads to the consideration of an infinite-horizon discounted robust control value of the following form:
\begin{equation}\label{eqn:r_ctrl_val_intro}
v^*(x,\Pi,\Gamma):= \sup_{\pi\in\Pi}\inf_{\gamma\in \Gamma} E_{x}^{\pi,\gamma} \sum_{t=1}^\infty \alpha^t r(X_t,A_t,W_t),
\end{equation}
subject to the system dynamics \eqref{eqn:state_dynamics}. 
Here, $x$ represents the initial state of the system, $\Pi$ and $\Gamma$ are the controller's and adversary's policy classes, respectively, and $\alpha\in(0,1)$ is the discount factor. Notably, the adversary's policy class $\Gamma$ permits the use of time- and history-dependent strategies. Hence, the formulation naturally accounts for history and temporal dependence, nonstationarity, and targeted attacks that arise in practice, while quantifying worst-case performance under such misspecification.

To enhance the modeling flexibility of the framework and tailor it to specific environments, we introduce two adversary types: current-action-aware (CAA) and current-action-unaware (CAU). In the CAA model, the adversary observes the realized action before selecting the perturbation, whereas in the CAU model, it does not. This distinction enables the modeler to calibrate how strategically the environment may react, thereby improving realism and resolution in managerial applications. We also provide managerial insights on how these two adversary models differ, guiding practitioners in selecting the appropriate specification for their setting; see Section \ref{sec:caa_vs_cau}. In particular:

\begin{itemize}
    \item Use CAU adversaries when the environment’s same-period noise are effectively exogenous to the manager’s action—typical in queueing, replenishment, or other settings where today’s arrivals or demand do not react instantly.
    \item Switch to a CAA model once the action can trigger an immediate environmental response (e.g., admission-control tweaks, flash promotions, trades with market impact). In these settings, a CAA adversary can be employed to capture the action-driven disturbances.
\end{itemize}

Importantly, the DRSC framework under both types of adversaries retains the core structural properties that make the classical MDP practical. We establish that the value function \eqref{eqn:r_ctrl_val_intro} satisfies a robust Bellman equation analogous to the nominal case, and hence stationary Markov policies are optimal. This is particularly relevant for managerial applications, as stationary policies are easy to communicate, straightforward to implement, and widely adopted in practice. Moreover, stationary optimality and the Bellman equation characterization not only ensure practicality but also support statistical learning and algorithmic computation, as both can build directly on the same dynamic-programming foundation as in the classical MDP framework.

Statistically, we show that, given knowledge of the transition dynamics and reward functions, together with data on the environment input sequence, optimal robust decisions can be learned efficiently under our DRSC framework. Specifically, we analyze both CAA and CAU adversaries constrained by either Wasserstein distance or \(f_k\)-divergence ambiguity sets, which represent the two principal classes of distributional ambiguity in distributionally robust optimization. Wasserstein ambiguity sets capture model errors at the outcome level, while \(f_k\)-divergence sets hedge against misspecifications in the likelihood of possible outcomes. Under suitable assumptions, we establish that the estimation error converges at parametric rates and, perhaps surprisingly, does not suffer from the curse of dimensionality—a phenomenon commonly associated with nonparametric estimation problems. We further provide matching lower bounds on these convergence rates, demonstrating the optimality of our results. A summary of these findings is presented in Table~\ref{tab:summary_of_results}.

\begin{table}[ht]
\begin{center}
\caption{Results on the statistical complexity of learning DRSC ($\widetilde\Theta$ suppress a gap of $\sqrt{\log n}$). }
\label{tab:summary_of_results}
\vspace{0.1 in}
\begin{tabular}{llll}
\toprule
Ambiguity  set                      & Adversary type & Action space & Complexity                      \\
\midrule
\multirow{2}{*}{Wasserstein distance} & CAA  & Continuum    & \multirow{2}{*}{$\Theta\crbk{n^{-1/2}}$}           \\
                                      & CAU  & Finite       &                                                       \\
\midrule 
\multirow{2}{*}{$f_k$-divergence}     & CAA  & Continuum    & \multirow{2}{*}{$\widetilde\Theta\crbk{n^{-\frac{1}{k'\vee 2}}}$}\\
                                      & CAU  & Finite      &                       \\
\bottomrule
\end{tabular}
\end{center}
\end{table}

Computationally, as discussed earlier, the optimal robust value function and policy satisfy the robust Bellman equation. This structure enables efficient implementation by leveraging existing algorithms for classical MDPs and reinforcement learning (RL) with neural-network function approximation. We show that standard RL pipelines can be extended to solve for optimal DRSC. In particular, we propose actor–critic methods that employ separate neural networks to parameterize the robust value function and the policy: the critic updates the robust value estimates through Bellman error minimization, while the actor improves the policy based on the critic’s feedback. This design yields robust policies with a comparable computational footprint to their non-robust counterparts.

In the two sets of numerical experiments, we implement the proposed actor-critic methods in both inventory control and portfolio management. We note that our methods converge quickly in practice to the robust value function and optimal policy. With the converged robust value and policy, we find that the CAA and CAU models have different performances, and are both better than the non-robust optimal policy. Based on the numerical results, we suggest that the randomization of CAU policy is suitable to deal with auto-correlation in the time series, and the CAA policy is generally more conservative than its CAU counterpart.


\subsection{Literature Review}
\label{sec:literature}
Distributionally robust stochastic control is not a new concept in the literature.  \citet{Yang2021Wasserstein_dr_control} investigate a setting aligned with our current-action-aware formulation, employing a Wasserstein uncertainty set. In contrast, \citet{petersen2000minimax} consider an uncertainty set based on Kullback–Leibler divergence.
For linear systems, where $f$ is linear on $X_t$,$A_t$, and $W_t$, distributionally robust stochastic control has been explored by several authors 
\citep{taskesen2024distributionally,tacskesen2025optimality,kim2023distributional,han2023distributionally,kotsalis2021convex}. Nonetheless, existing research predominantly focuses on characterizations of the optimal policies or the development of tractable optimization methods. In our study, we address the statistical complexity associated with the learning problem.

Our work is also closely related to the literature on DRMDPs and distributionally robust reinforcement learning (DRRL). Various formulations are explored in \citet{iyengar2005robust, nilim2005robust, le_tallec2007robustMDP, xu2010distributionally, wiesemann2013robust, wang2023foundation, Goyal2023Beyond_Rectangularity, li_shapiro2023rectangularity}. Concurrent with this manuscript, \cite{shapiro2024drsc} also explores DRSC. However, the focus of the two papers differs significantly: whereas \cite{shapiro2024drsc} emphasizes formulation aspects of DRSC, our work formulates and investigates statistical properties associated with optimal policy learning within a DRSC framework.

\par Statistical complexities for DRRL problems with finite state and action spaces have been developed in recent literature. Most studies focus on the SA-rectangular setting, which aligns with the behavior of our CAA adversary \citep{zhou21, Panaganti2021, yang2021, ShiChi2022, xu2023improved, shi2023curious_price, blanchet2023double_pessimism_drrl, liu22DRQ, Wang2023MLMCDRQL, wang2023VRDRQL, yang2023avoiding}, while several recent results also address the S-rectangular setting, which is related to our CAU formulation \citep{yang2021, Clavier2024Srec, li2025Srec}.


\par However, the picture is very different in the continuous state space setting. In fact, \citep{chen2019information} show that even in finite MDPs, if one does not impose mild distribution-shift (concentratability) or completeness assumptions, no batch RL algorithm can achieve polynomial sample complexity — hence efficient RL is impossible without such restrictions. These considerations become more stringent in the continuous setting and this challenge also applies to DRRL.


Our formulation of DRSC differs from DRRL as we assume a known state recursion form $f$ driven by an unknown random variable $W$, whereas in DRRL, the full transition probabilities need to be learned. This difference leads to a significant difference in sample complexities. We derive minimax optimal sample complexities for uniformly estimating the robust value function with parametric convergence rates even in continuous state and action spaces.

\subsection*{Remarks on Paper Organizations}

The remainder of this paper is structured as follows: Section \ref{sec:DPP} outlines the formulations of the DRSC problems and the corresponding dynamic programming principles, with a focus on both CAA and CAU adversaries. Sections \ref{sec:UB} and \ref{sec:LB} present the upper and lower bounds of the sample complexities, respectively. Section \ref{sec:algoDesign} presents the algorithm design for solving the robust value and policy. Finally, Section \ref{sec:experiment} presents two applications within our framework to demonstrate its effectiveness in modeling real-world problems. 

Some preliminary theoretical results have appeared in a conference version of this
paper. However, the current version contains substantial  additional results that go beyond statistical
analysis: it offers a more streamlined motivation and clearer managerial insights, extends the model
expressiveness by allowing the reward function to depend on the environmental noise sequence,
and provides new theoretical analyses, algorithmic developments, and numerical experiments on
both synthetic and real data.

\section{Distributionally Robust Stochastic Control Framework}\label{sec:DPP}

In this section, we provide a self-consistent introduction to our formulation of the distributionally robust stochastic optimal control problem and its corresponding dynamic programming theory. The fully rigorous construction is provided in Appendix \ref{a_sec:DRSC_formulation}. Our focus here is on the infinite horizon discounted reward setting. It should be noted that a DRSC formulation for finite-horizon systems naturally arises from the same principles. 
\par We consider Polish (i.e. complete separable metric spaces) state, action, and noise measurable spaces $(\X,\cX),(\A,\cA),(\W,\cW)$ equipped with the Borel $\sigma$-fields generated by open sets. Let $\cP(\cW)$ and $\cP(\cA)$ denote the set of probability measures on the action and noise spaces, endowed with the topology of weak convergence. As motivated in the introduction, we consider a known state dynamic function $f:\X\times\A\times\W\ra\X$ given in \eqref{eqn:state_dynamics}, a known reward function $r:\X\times\A\times\W \ra \R_+$, and a discount factor $\alpha\in(0,1)$.

\subsection{Controller and Adversary Policies}
\par Let $\cQ\subset\cP(\cA)$ and $\cP\subset\cP(\cW)$ be arbitrary Borel measurable subsets. Here, $\cQ$ and $\cP$ represent the admissible decision sets of the controller and the adversary, respectively. Based on $\cQ$ and $\cP$, we construct admissible policy classes $\Pi(\cQ)$ and $\Gamma(\cP)$ of the controller and adversary, respectively. As we will rigorously develop in the Appendix \ref{a_sec:DRSC_formulation}, the admissible control policy class $\Pi(\cQ)$ of the controller will always be a subset of the history-dependent $\cQ$-constrained policy class:
$$\Pi(\cQ)\subset\PiH(\cQ):= \set{\pi = (\pi_0,\pi_1,\ds):\pi_t(da|h_t) \in\cQ,\forall h_t = (x_0,a_0,\ds,a_{t-1},x_t)}.$$ 
We will suppress the dependence on $\cQ$ when the controller's action set is clear from the context. Intuitively, the controller decides a sequence of the conditional distribution of the current action $A_t$ given the history until the last visible state $x_t$. 

Two controller policy classes of particular interest are the history-dependent policies $\PiH$ and the stationary Markov policies $\PiS \subset \PiH$, where 
\[
\PiS := \set{\pi \in \PiH : \pi_t(da \mid h_t) = \pi(da \mid x_t), \; \forall t \geq 0,\; h_t = (x_0,a_0,\ldots,x_t)}.
\]
Hence, $\PiS$ can be identified with measurable functions $\X\ra \cQ$; i.e. for each measureable $\eta:\X\ra \cQ$, $\pi = (\eta,\eta,\eta,\ds) \in\PiS$. We develop dynamic programming theories to identify modeling environments in which the best robust policy within $\PiH$ achieves the same robust control value \eqref{eqn:r_ctrl_val_intro} as the best robust policy in $\PiS$.

\par Similarly, adversary's admissible policy class $\Gamma(\cP)$ is a subset 
$$\Gamma(\cP)\subset\GammaH(\cP) := \set{\gamma = (\gamma_0,\gamma_1,\ds):\gamma_t(dw|g_t)\in\cP, \forall g_t = (x_0,a_0,\ds,x_t,a_t)}.$$
Again, we suppress the dependence on $\cP$ when the adversary's action set is clear. The adversarial policy $\gamma\in\Gamma$ determines the conditional distribution of $W_t$ given the history until the last visible state action pair $x_t,a_t$, for each and every $t\ge 0$. Like in the controller case, the stationary Markov adversary policy class is 
$$\GammaS := \set{\gamma \in\GammaH:\gamma_t(dw|g_t) = \gamma(dw|x_t,a_t) , \forall t\geq 0, g_t = (x_0,a_0,\ds,x_t,a_t)},$$
which can be identified with measurable mappings $\X\times\A\ra\cP$. We remark that, unlike on the controller side, in Section \ref{sec:caa_vs_cau} we further distinguish a special type of adversarial behavior, termed current-action-unawareness, which gives rise to a more restricted policy class $\barGammaH \subset \GammaH$ and its stationary counterpart $\barGammaS \subset \GammaS$.

\subsection{System Dynamics and Optimal Robust Control Value}\label{sec:ctrl_adv_dynamics}

\par For any given pair of controller and adversarial policy $(\pi,\gamma)\in\PiH\times\GammaH$ with and an initial state $x\in\X$, the distribution of the state and action process $(X,A)$ is uniquely determined, see \eqref{eqn:def_prob_meas_from_pi_gamma}, and we denote the expectation under this distribution as $E_x^{\pi,\gamma}$. To clarify, we provide an intuitive description of the stochastic system dynamics under $E_x^{\pi,\gamma}$. Let $H_t = (X_0,A_0,\ds ,A_{t-1},X_t)$ denote the history at time $t$. Given controller policy $\pi=(\pi_0,\pi_1,\ldots)$ and an adversary policy $\gamma=(\gamma_0,\gamma_1,\ldots)$, the expectation $E^{\pi,\gamma}_x$ is equivalent to the path measure generated by the following dynamic procedure. Initialize with $t = 0$, $X_t = x$. 
\begin{enumerate}
  \item Draw controller's action $A_t \sim \pi_t(\cdot | H_t).$
  \item Draw the adversarial environment input $W_t \sim \gamma_t(\cdot | H_t,A_t).$
  \item Update the state via the transition dynamics $X_{t+1}=f(X_t, A_t, W_t)$, 
  set $t \leftarrow t+1$, and repeat.
\end{enumerate}
This procedure determines the distributional properties of the entire $\set{X_t,A_t,W_t:t\geq 0}$ sequence.

Then, given controller's and adversary's action sets $\cQ$ and $\cP$, we define the optimal robust control value function given any controller's and adversarial policy classes $\Pi\subset\PiH$ and $\Gamma\subset\GammaH$ as
\begin{equation}\label{eqn:simp:DRSC_value_func}
v^*(x,\Pi,\Gamma):= \sup_{\pi\in\Pi}\inf_{\gamma\in \Gamma} E_{x}^{\pi,\gamma} \sum_{t=1}^\infty\alpha^tr(X_t,A_t,W_t).
\end{equation}

From the system dynamics and objective, we observe that the DRSC framework induces robustness by modeling the environment as if it were controlled by an adversary capable of perturbing the input sequence. The decision maker thus optimizes not against a single, possibly misspecified model, but against all models within a prescribed ambiguity set. This adversarial viewpoint stress-tests policies across a broad range of dynamics, rather than tailoring them to a nominal i.i.d.\ specification that may fail in practice. Thus, by planning for the most adverse disturbances, the decision maker builds a margin of safety into the policy, safeguarding against model misspecification and enhancing reliability and resilience in deployment.

\section{Adversarial Models and Dynamic Programming}

In this section, we introduce two adversarial models—current-action-aware (CAA) and current-action-unaware (CAU)—highlight their modeling implications, and develop their respective dynamic programming theories. To set the stage, we first lay out the regularity assumptions under which the optimal robust control values can be characterized through their dynamic programming equations, i.e., robust Bellman equations.

\subsection{Current-Action-Awareness versus Unawareness}\label{sec:caa_vs_cau}
We are now ready to formally define the CAA and CAU adversaries, clarify the modeling implications, and discuss the corresponding dynamic programming theories that characterize optimal robust control under these two settings.

\subsubsection*{Current-Action-Aware Adversary: }

Observe that, by the construction in Section \ref{sec:ctrl_adv_dynamics}, a general adversary policy $\pi\in\GammaH$ induces different conditional distributions of $W_t$ given different realizations of $A_t$. Therefore, one can think about the policy class $\GammaH$ as one in which the adversary is aware of the current action. Hence, we call $\GammaH$ the current-action-aware adversary policy class.

Since an adversarial policy $\pi \in \PiH$ can use the realization of the controller’s current action to harm performance, the CAA adversary provides a natural model for settings where ambiguity or misspecification of the environment input arises from the system’s reaction to the control action—behavior that is not captured under an i.i.d. input model. This situation can occur, for example, when the sequence ${W_t}$ is endogenously coupled with the current action.

An illustrative management example is supply chain control, where $W_t$ represents product demand and the current action $A_t$ encodes multiple managerial decisions, including but not limited to pricing. In this case, it is reasonable to expect that demand may systematically respond, potentially in an adversarial manner from the manager’s perspective, to those pricing decisions. See Example \ref{exmpl:inventory} for a concrete discussion. 

A CAA model is also suitable in high-risk environments where the system’s response is endogenously tied to the current action, even if not explicitly adversarial. Portfolio management (cf. Example \ref{exmpl:portfolio}) provides such a case: although it may be unclear whether buying or selling an asset triggers cooperative or adverse market reactions, adopting a CAA perspective enables the decision maker to remain risk-aware and volatility-averse.

\par For CAA adversaries, we consider the following distributionally robust Bellman equation. 
\begin{definition}[CAA Robust Bellman Equation]
Given the controller’s and the adversary’s action sets $\cQ$ and $\cP$, we say that $u^*$ is a solution to the current-action-aware robust Bellman equation if
\begin{equation}\label{eqn:def_caa_DR_bellman_eqn_action_int}
\begin{aligned}
u^*(x) &= \sup_{\phi\in\cQ} \int_{\A} \inf_{\psi\in \cP}\int_{\W} r(x,a,w) + \alpha u^*(f(x,a,w))\psi(dw)\phi(da). 
\end{aligned}
\end{equation}
Notice that if $\cQ \supset \set{\delta_{\set{a}}: a\in \A}$, then the supremum over $\cQ$ is equivalent to taking the supremum over the set of point-mass actions $\set{\delta_{\set{a}}:a\in \A}$. Consequently, in this setting, the CAA robust Bellman equation can be written as
\begin{equation}\label{eqn:def_action_aware_DR_bellman_eqn} u^*(x)=\cT(u^*)(x) :=\sup_{a\in\A} \inf_{\psi\in \cP}\int_{\W}r(x,a,w) + \alpha u^*(f(x,a,w))\psi(dw) \end{equation}
where $\cT$ denotes the CAA robust Bellman operator.
\end{definition}
\par We will show in Theorem \ref{thm:main_body_dpp} that, under appropriate regularity conditions, the CAA Bellman equation \eqref{eqn:def_caa_DR_bellman_eqn_action_int} has a unique bounded solution $u^*$. Moreover, the optimal robust control value in \eqref{eqn:simp:DRSC_value_func} is equal to $u^*$. 

Note that, in the CAA equation \eqref{eqn:def_action_aware_DR_bellman_eqn}, the infimum operator—representing the adversary’s decision—appears inside the action integral. This reflects an adversary that chooses its decision after observing the current realized action, hence action-awareness.

\subsubsection*{Current-Action-Unaware Adversary: }
Although CAA adversaries yield natural distributionally robust control models in many applications, there also exist important managerial settings where the environment inputs are essentially exogenous, evolving according to dynamics that are largely decoupled from the state–action sequence. Such cases motivate the formulation and use of the current-action-unaware adversary policy classes.

\par Concretely, an adversarial decision $\bar \gamma_t$ at time $t$ is said to be \emph{current-action-unaware} if, for any history $h_t = (x_0,a_0,\ds,x_{t})$, we have 
\[
\bar \gamma_{t}(dw \mid h_t,a) = \bar \gamma_t (dw \mid h_t,a'), \quad \forall a,a'\in\A.
\] 
Recall the adversarial dynamics described in Section \ref{sec:ctrl_adv_dynamics}, the CAU structure implies that the conditional distribution of $W_t$ given the history is independent of the current action $A_t$.

We then define the history-dependent CAU adversary policy class as 
\[
\barGammaH := \set{\bar\gamma = (\bar\gamma_0,\bar\gamma_1,\ds)\in\GammaH:\bar\gamma_t \text{ is current-action-unaware } \forall t\geq 0}. 
\] 
The stationary version, denoted by $\barGammaS$, is defined analogously. Throughout the paper, we use the overline notation to indicate settings with CAU adversaries.

We note that, compared to the robust MDP literature, a CAU policy class can be viewed as a special case of the S-rectangular policy class; see \citet{wiesemann2013robust,wang2023foundation} and the references therein. However, in addition to S-rectangularity, the CAU structure implicitly leverages the system dynamics induced by the transition function $f$. This connection offers greater interpretability and practical relevance in managerial settings where the stochastic environment evolves fully exogenously, or more broadly, independently of the controller’s immediate action. 

To build theoretical intuition, consider a simple case where the environment input sequence is close to, but not exactly, i.i.d. Instead, it follows an AR(1) process $W_{t} = \delta W_{t-1} + D_t$, where $\{D_t\}$ is an i.i.d.\ sequence and $\delta$ is a correlation parameter. In this case, $W_t$ evolves exogenously, fully decoupled from the state–action process.  

From the transition dynamics $X_t = f(X_{t-1},A_{t-1},W_{t-1})$, we see that $W_{t-1}$ and $X_t$ are dependent, and since $X_t$ depends on the entire history $H_t$, it follows that $W_t$ is also dependent on $H_t$. On the other hand, conditioned on $H_t$, the next action is generated by the controller’s decision rule $A_t \sim \pi_t(\cdot \mid H_t)$ and is therefore independent of both $W_{t-1}$ and $D_t$. Hence, given $H_t$, $W_t$ is conditionally independent of $A_t$. These structural properties are naturally captured by a history-dependent CAU adversary. Thus, in settings where the environment sequence follows an exogenous dynamics, the CAU formulation is the more appropriate modeling choice.

For example, in the robust control of a make-to-order manufacturing system (cf.\ Example \ref{exmpl:service_system}), it is reasonable to assume that the inter-arrival times follow an exogenous process, in which case a CAU model is appropriate. By contrast, if one adopts a CAA model and allows adversarially perturbed inter-arrival times to depend directly on the controlled service rate, the resulting policy may become overly conservative, thereby reducing value in deployment environments where arrivals are in fact independent of the service assignment. A similar argument applies in inventory management, where the order quantity $A_t$ does not directly influence the product demand $W_t$ on day $t$.

\par For CAU adversaries, we propose the following distributionally robust Bellman equation. 
\begin{definition}[CAU Robust Bellman Equation]
Given the controller’s and the adversary’s action sets $\cQ$ and $\cP$, we say that $\bar u^*$ is a solution to the current-action-unaware robust Bellman equation if
\begin{equation}\label{eqn:def_action_unaware_DR_bellman_eqn}
\bar u^*(x) = \overline\cT(\bar u^*)(x):= \sup_{\phi\in\cQ}\inf_{\psi\in \cP} \int_{\A\times\W} r(x,a,w) + \alpha \bar u^*(f(x,a,w))\phi\times\psi(da,dw) 
\end{equation}
where $\overline\cT$ is the CAU robust Bellman operator. 
\end{definition}
The existence and uniqueness of the solution, together with the corresponding dynamic programming principle, are established in Theorem \ref{thm:main_body_dpp}.

Notice that, compared to the CAA equation \eqref{eqn:def_caa_DR_bellman_eqn_action_int}, the CAU equation places the infimum operator—corresponding to the adversary’s decision—outside of the action integral. This structure reflects action-unawareness, as the adversary must use the same $\psi \in \cP$ for all actions $a \in \A$.

\subsection{Existence and Uniqueness of Solution and Dynamic Programming}

With these formulations and adversarial modeling considerations in place, we now establish the existence and uniqueness of solutions to the proposed CAA and CAU robust Bellman equations, along with their corresponding dynamic programming principles.  

Throughout the remainder of the paper, we use $\norm{\cdot}$ to denote the supremum norm and define $\beta := (1-\alpha)^{-1}$ for notational convenience. As is standard in the stochastic control literature, we impose regularity conditions on the reward function $r$ and the state transition function $f$. These conditions guarantee that the Bellman equations admit well-defined and regular solutions.

\begin{assumption}\label{assump:cond_for_DPP}
Assume the following conditions on the measurable reward function $r:\X\times\A\times \W\ra\R$, transition function $f:\X\times\A\times \W\ra\X$, and controller action set $\cQ\subset\cP(\cA)$:
\begin{itemize}
    \item $r$ is nonnegative and uniformly bounded by $r_\vee$; i.e., $0 \leq r(x,a,w)\leq r_\vee ,\forall x,a,w$.
    \item $r$ and $f$ are uniformly continuous.
    \item The controller is able to any choose deterministic actions; i.e. $\cQ \supset \set{\delta_{\set{a}}: a\in \A}$. 
\end{itemize}
\end{assumption}

\begin{remark}
For the purpose of establishing a dynamic programming theory, these conditions can in fact be weakened: boundedness may be replaced by growth conditions, and uniform continuity can be relaxed to weaker forms of semi-continuity (see, e.g., \citet{gonzalez2002minimax}). 
\end{remark}

With this said, one of the main goals of this paper is to analyze the statistical complexity and develop algorithms for learning the optimal robust policy. For these purposes, we impose stronger assumptions—such as uniform Lipschitz continuity, stated later in Assumptions \ref{assump:aware_lip_val_bounded_sps} and \ref{assump:unaware_lip_val_bounded_sps}—which allow us to control estimation error and establish sharp parametric learning rates. While a more general dynamic programming theory would certainly be of theoretical value, we adopt these more restrictive but easier-to-communicate assumptions to make the framework accessible to a broader audience.

\begin{theorem}\label{thm:main_body_dpp} Suppose Assumption \ref{assump:cond_for_DPP} is in force. Then the following holds
\begin{itemize}
    \item The robust Bellman equations \eqref{eqn:def_action_aware_DR_bellman_eqn} and \eqref{eqn:def_action_unaware_DR_bellman_eqn} have unique fixed points $u^*$ and $\bar u^*$, respectively within the class of uniformly continuous and non-negative functions $\X\ra\R_+$. 
    \item The solutions are also bounded:  $\norm{u^*},\norm{\bar u ^*}\leq \beta r_\vee.$
    \item In the CAA case, $u^* = v^*(\cd,\Pi,\Gamma)$ for any combination of $\Pi = \PiH,\PiS$ and $\Gamma = \GammaH,\GammaS$. 
    \item In the CAU case, $\bar u^* = v^*(\cd,\PiH,\barGammaH) = v^*(\cd,\PiS,\barGammaH) = v^*(\cd,\PiS,\barGammaS)$.
\end{itemize}

\end{theorem}
\begin{remark}
    Note that $v^*(\cdot,\PiH,\GammaH) = v^*(\cdot,\PiS,\GammaH)$ and $v^*(\cdot,\PiH,\barGammaH) = v^*(\cdot,\PiS,\barGammaH)$ imply that, to achieve optimal robust control, it suffices for the decision maker to restrict attention to stationary Markov policies. 
    We also highlight an important difference between the dynamic programming results in the CAA and CAU cases under the controller-adversary pairing $(\PiH,\barGammaS)$. In the CAU setting, it is generally the case that $\bar u^* \neq v^*(\cdot,\PiH,\barGammaS)$; counterexamples can be constructed following those in \citet{wang2023foundation}. However, under additional convexity and compactness assumptions, one can obtain $\bar u^* = v^*(\cdot,\PiH,\barGammaS)$; see also \citet{wang2023foundation}.
\end{remark}

With Theorem \ref{thm:main_body_dpp} in place, we conclude that learning the optimal robust control value and thus the corresponding robust control policy reduces to finding accurate approximations to the solutions of \eqref{eqn:def_action_aware_DR_bellman_eqn} and \eqref{eqn:def_action_unaware_DR_bellman_eqn} in the CAA and CAU settings, respectively.

\subsection{Modeling Examples}
\label{subsec:modeling_examples}

We illustrate the practicality of adversarial models by three typical managerial examples. In each setting, we first formulate the stochastic model, and then discuss the choice of CAA or CAU adversarial models.

\begin{example}[Service or manufacturing systems]\label{exmpl:service_system}
Consider a make-to-order single-server system. Let $X_n$ be the waiting time of job $n$ just after it arrives. Each job requires one unit of work. When job $n$ arrives, we choose a service rate $A_n>0$ (via staffing, machine power, or utilization), so its service time is $S_n=1/A_n$. Let $W_n$ be the inter-arrival time between job $n$ and job $n{+}1$. The waiting-time recursion is
\[
X_{n+1} = f(X_n,A_n,W_n) := \crbk{X_n + \tfrac{1}{A_n} - W_n}_+.
\]
A simple one-step reward penalizes delay and effort,
\[
r(X_n,A_n,W_n) := -\,h\,\crbk{X_n + \tfrac{1}{A_n} - W_n}_+ - c(A_n),
\]
with $h>0$ representing the cost per unit of waiting time, and $c$ is the cost of using service rate $A_n$.

\textbf{CAA vs.\ CAU: }In most service systems, the arrival process is determined upstream (marketing, demand generation, external customers), hence typically exogenous. Inter-arrival $W_n$ does not react to the rate chosen for the job, so CAU is the natural model: the adversary can use history to shape $W_n$ but cannot condition on $A_n$. Choose CAA only if the within-period action plausibly changes the same-period arrivals—for example, if the manager simultaneously adjusts admission control or releases time-sensitive promotions that immediately alter arrivals—otherwise CAA could lead to needlessly conservative controls.
\end{example}

\begin{example}[Inventory control]\label{exmpl:inventory}
Let $X_t\in\R_+$ be on-hand inventory at the start of day $t$. In the morning we deliver $A_t\in\R_+$ units, so on-hand becomes $X_t+A_t$. Let $W_t\in\R_+$ be the demand realized during day $t$. With no backlogging,
\[
X_{t+1} = f(X_t,A_t,W_t) := \crbk{X_t + A_t - W_t}_+.
\]
A standard daily reward is profit net of costs and penalties,
\[
r(X_t, A_t, W_t) = p\min (X_t+A_t, W_t) -c A_t -h X_t,
\]
where $p$ is unit margin, $c$ the order/transfer cost, $h$ holding cost.

\textbf{CAA vs.\ CAU:} In this simple form, the action is a pure replenishment decision, i.e., deliveries from upstream logistics with demand realized later that day. Then, the same‑day demand does not react to $A_t$, so CAU is appropriate. One can consider using CAA when the action includes levers that change demand within the day (price, promotions, availability signals that prompt immediate purchases). In that case, allowing the adversary to condition on $A_t$ captures fast, action‑dependent demand shifts; otherwise, the model can underestimates risk from demand surges triggered by the decision.
\end{example}

\begin{example}[Portfolio management]\label{exmpl:portfolio}
We manage $m$ assets. Let $X_t\in\R^m$ be the vector of dollar holdings at the start of period $t$ ($X_{t,i} > 0$ and $X_{t,i} < 0$ mean a long position and a short position in asset $i$, respectively) and let $A_t\in\R^m$ be resulting percentage of the $m$ assets after trading at $t$ ($A_{t,i} > 1$ or $0 \le A_{t,i} < 1$ means buying or selling asset $i$ at the beginning of time period $t$, respectively). Let $W_t=\mathrm{diag}(R_t)\in\R^{m\times m}$ be the diagonal matrix of gross returns from $t$ to $t{+}1$. 

For simplicity, we assume that there is no contribution of capital. However, one can consume the portfolio by having a total sale that is higher than the total purchase; i.e. $(A_t-\bd{1})^\top X_t \leq 0$ where $\bd{1}\in\R^m$ is the column vector of all 1's. Therefore, the consumption is $C_t = \bd{1}^\top X_t - A_t^\top X_t\ge 0$, which is the cash amount taken out from the portfolio. As such, the self-financing constraint (in the absence of any transaction costs) becomes $ (A_t - \bd{1})^\top X_t + C_t = 0$. 

At the beginning of the next period, we have 
\[
X_{t+1} = W_{t} (A_t \odot X_t).
\]

\par We assume that the decision maker is endowed with a utility function $u: \mathbb{R}_+ \rightarrow \R$, which is concave and non-decreasing. Then, taking action $A_t$ will incur an instantaneous reward utility 
\[
r(X_t,A_t,W_t) = u(\bd{1}^\top X_t - A_t^\top X_t).
\]

\textbf{CAA vs.\ CAU: } If trades move prices through market impact, the return law effectively reacts to the current trade. Model this with CAA so the adversary can condition $W_t$ on $A_t$ (temporary impact, slippage, liquidity dry‑ups). If the trader is small relative to the market or executes passively in deep venues, it is reasonable to treat returns as exogenous to the current trade. Then CAU captures serial dependence and regime change without over‑penalizing actions. In short: use CAA to reflect immediate action‑driven risk; use CAU when returns are driven by external factors and impact is negligible.
\end{example}

\section{Learning Optimal DRSC: Statistical Complexity Upper Bounds}\label{sec:UB}

Beyond modeling expressiveness, a key requirement of the DRSC framework is statistical efficiency: given a reasonably large dataset, we should be able to infer near-optimal policies with reliable fidelity. To provide useful statistical insights and sharp convergence guarantees, it is necessary to tailor the complexity analysis to the specific form of the ambiguity set, since allowing completely general ambiguity sets renders the problem statistically intractable in a minimax sense. In this paper, we therefore focus on two widely used ambiguity models—Wasserstein distance and $f_k$-divergence—which capture distinct facets of distributional uncertainty.

When deploying DRSC models, different applications motivate different notions of robustness. At a high level, the design of ambiguity sets for the environment input depends on which aspect of misspecification is most concerning. In some cases, the focus is on outcomes: if realized states or environmental inputs may be perturbed within a small range—due to measurement noise, regime shifts, or model misspecification in the dynamics—it is natural to adopt Wasserstein-type ambiguity sets, which capture closeness in the geometry of the input space \citep{mohajerin2018data}. In other cases, the concern lies in likelihoods: the support of outcomes may be trusted, but their relative probabilities may be misspecified, for instance due to sampling bias or confounding. Here, likelihood-based sets, such as those defined by $f_k$-divergences, are more suitable because they measure robustness in terms of reweighting likelihood ratios \citep{ben2013robust}. These two modeling perspectives—outcome perturbations versus likelihood perturbations—motivate the use of Wasserstein and $f_k$-divergence ambiguity sets in our framework. 

At the same time, Wasserstein or $f_k$ ambiguity models enjoy favorable statistical or computational properties. From the distributionally robust optimization literature, Wasserstein balls often yield strong generalization guarantees \citep{gao2017distributionally}, while $f_k$-divergence balls lead to closed-form dual formulations and efficient optimization \citep{duchi2019distributionally}.

We now formally introduce Wasserstein distance and $f_k$-divergence ambiguity sets (Definitions \ref{def:Wd_set} and \ref{def:f_k_div_set}). In both cases, we fix a reference measure $\mu_0 \in \cP(\cW)$ as the center of the ambiguity set, assumed to be accessible from sampling.

\begin{definition}[Wasserstein distance ambiguity sets]\label{def:Wd_set}
Let $c:\W\times\W \ra\R_+$ s.t. $c(w,w) = 0$ for all $w\in\W$. The Wasserstein distance for $\mu,\nu\in\cP(\cW)$ with transportation cost function $c$ is defined as 
$$W_c(\mu,\nu):=\inf_{\xi\in\Xi(\mu,\nu)}\int_{\W\times\W}c d\xi, $$
where $\Xi(\mu,\nu)$ is the set of probability measures on $\cW\times\cW$ s.t. $\xi(\cd,\W) = \mu$ and $\xi(\W,\cd) = \nu$ for all $\xi\in\Xi(\mu,\nu)$. The Wasserstein distance constrained ambiguity set of adversarial decisions with cost $c$, transport budget $\delta$ and center $\mu_0$ is 
$\cP = \set{\mu:W_c(\mu,\mu_0)\leq \delta}$. 
\end{definition}
\begin{definition}[$f_k$-divergence ambiguity sets]\label{def:f_k_div_set}
    For $k > 1$ and probability measures $\mu\ll\mu_0$ in $\cP(\cW)$, let 
    \[\begin{aligned}
    &f_k(t):=\frac{t^k - kt+k-1}{k(k-1)}, \\
    &D_{f_k}(\mu\|\mu_0) := \int_{\W} f_k\crbk{\frac{d\mu}{d\mu_0}}d\mu_0. 
    \end{aligned}
    \] 
    Then, the $f_k$-divergence constrained  ambiguity set of adversarial decisions with center $\mu_0$ and radius $\delta$ is 
    $\cP = \set{\mu\ll\mu_0: D_{f_k}(\mu\|\mu_0)\leq \delta}$. 
\end{definition}

\subsection{Data, Learning Objective, and Estimator}
To achieve robust and data-driven decision-making, we seek to learn system characteristics and robust decisions from data whose distributional properties may not perfectly align with those in the deployment environment. We assume access to a data set that may be collected either from a simulator of the real system of interest or directly from operations during a data collection period. For analytical tractability, we model this data set as $n$ i.i.d. samples $\{W_i : i=1,\ldots,n\}$ with $W_i \sim \mu_0$.  

Our learning objective is to estimate the optimal robust value defined in \eqref{eqn:def_action_aware_DR_bellman_eqn} and \eqref{eqn:def_action_unaware_DR_bellman_eqn} across the entire state space $\X$. Specifically, we aim to construct an estimator $\hat u$ of the optimal robust value functions (here denoted by $u$) such that the uniform error  
\[
\|u - \hat u\| = \sup_{x\in\X} |u(x)-\hat u(x)|
\]  
remains small. This objective requires learning an infinite-dimensional object while ensuring uniform accuracy across the entire state space.  

To address this challenge, we adopt the empirical robust Bellman estimator (equivalently, the plug-in estimator in this context), constructed using the empirical measure–centered version of the ambiguity sets introduced below.

For both the Wasserstein distance and  $f_k$-divergence settings, we let $\widehat\cP$ do denote the $n$-sample empirical measure-centered version of the ambiguity sets; i.e. one that replaces $\mu_0$ with $\hat\mu$ where $$\hat\mu(\cd):=\frac{1}{n}\sum_{i=1}^n\1\set{W_i\in \cd}$$
for i.i.d. $W_i\sim\mu_0$. Further, we denote the CAA/CAU empirical Bellman operator as $\bd{T}$/$\overline{\bd{T}}$, where $\cP$ in equation
\eqref{eqn:def_action_aware_DR_bellman_eqn}/\eqref{eqn:def_action_unaware_DR_bellman_eqn} is replaced by $\widehat \cP$, i.e., 
\begin{align*}
&\bd{T}(u)(x):=\sup_{a\in\A} \inf_{\psi\in \widehat {\cP}}\int_{\W} r(x,a,w) + \alpha u(f(x,a,w))\psi(dw), 
\end{align*}
and 
\begin{align*}
&\overline{\bd{T}}(\bar u)(x):=\sup_{\phi\in\cQ}\inf_{\psi\in\widehat{\cP}} \int_{\A\times\W} r(x,a,w) + \alpha \bar u(f(x,a,w))\phi\times\psi(da,dw).
\end{align*}
Then, the empirical versions of  \eqref{eqn:def_action_aware_DR_bellman_eqn} and \eqref{eqn:def_action_unaware_DR_bellman_eqn} are $\bd{T}(u)(x)=u(x)$ and $\overline{\bd{T}}(\bar u)(x)=\bar u(x)$. 
\begin{proposition}\label{prop:bound_est_err_by_oprtr_gap} Suppose Assumption \ref{assump:cond_for_DPP} is in force. Let $u'$ and $\hat u$ be the solution to the population and empirical versions of \eqref{eqn:def_action_aware_DR_bellman_eqn} or \eqref{eqn:def_action_unaware_DR_bellman_eqn}; let $\cT'$ and $\bd{T}'$ denote the corresponding population and empirical Bellman operators. Then, the estimation error in uniform norm is upper bounded by 
$$\norm{\hat u - u'}\leq \beta \norm{\bd{T}' (u') - \cT' (u')}. $$
\end{proposition}

In the following two sections, we establish statistical complexity upper bounds for learning the optimal value uniformly under CAA and CAU adversary models. In both cases, we consider both the Wasserstein distance and $ f_k$-divergence-based adversarial ambiguity set. Our focus here is to obtain a tight convergence rate in $n$. 
\subsection{The Current-Action-Aware Case}\label{section:UB_aware}
We begin by clarifying notations and stating the assumptions under which our statistical analysis is carried out. For set $S\subset\R^d$, let $\diam(S):= \sup_{x,y\in S}|x-y|$, where $|\cd|$ denotes the Euclidean distance. For the CAA adversary case, we assume the following. 
\begin{assumption}\label{assump:aware_lip_val_bounded_sps}
    Assume the following conditions: 
    \begin{enumerate}
        \item  The spaces $\X\subset\R^{d_{\X}},\A\subset\R^{d_{\A}},\W\subset\R^{d_{\W}}$ are equipped with the Euclidean distance. The state and action spaces are bounded:  $ \diam(\A),\diam(\X)<\infty$.
        \item To simplify notation, we require Assumption \ref{assump:cond_for_DPP} to hold with $r_\vee = 1$.
        \item The mapping $(x,a)\ra r(x,a,\cd) + u^*(f(x,a,\cd))$ is uniform $L$-Lipschitz, i.e. $$|r(x,a,w)) - r(x',a',w))| + |u^*(f(x,a,w)) - u^*(f(x',a',w))|\leq L(|x-x'| +|a-a'|).$$
    \end{enumerate}
\end{assumption}

We remark that, at an intuitive level, the assumptions of bounded domains and Lipschitz continuity are natural for achieving the $\widetilde O(n^{-1/2})$ convergence rate in estimating the robust optimal value. Bounded domains rule out the necessity to learn the behavior of the value function at infinity, which would otherwise require an infinite amount of data. Uniform Lipschitz continuity, on the other hand, reflects the natural smoothness inherent in many physical and economic systems (see \cite{asadi2018lipschitz}, \cite{ni2019learning}). It prevents the value function from exhibiting unbounded oscillations or infinite total variation, phenomena that could significantly deteriorate statistical rates. Together, these conditions ensure that small perturbations in the empirical distribution translate into controlled deviations in the robust value, thereby enabling sharp statistical guarantees.

In the following Theorems \ref{thm:aware_Wd_sample_complexity} and \ref{thm:aware_fk_sample_complexity}, $u^*$ and $\hat u$ are solutions to \eqref{eqn:def_action_aware_DR_bellman_eqn} with corresponding Wasserstein and $f_k$ adversarial ambiguity sets $\cP$ and $\widehat\cP$ centered at $\mu_0$ and $\hat\mu$, respectively. 
\begin{theorem}[Wasserstein distance constrained CAA adversary]\label{thm:aware_Wd_sample_complexity} Suppose Assumption
 \ref{assump:aware_lip_val_bounded_sps} is in force. Also, assume that the cost is $c_\vee$-bounded; i.e. $\sup_{w,y\in\W}c(w,y)\leq c_\vee$.  Then, with $\cP = \set{\mu:W_{c}(\mu,\mu_0)\leq\delta}$, we have that
    $$\norm{\hat u - u^*}\leq  \crbk{32\sqrt{\pi} \beta D_\vee\sqrt{d_{\X}+d_{\A}+1}+ \sqrt{2}\beta^{3/2}\sqrt{\log\crbk{\frac{1}{\eta}}}}n^{-\frac{1}{2}}$$
 w.p. at least $1-\eta$. Here, $D_\vee:= L(\diam(\X) + \diam(\A)) + c_\vee\delta\inv \beta + 1 $. 
\end{theorem}
\begin{remark}
Employing the chaining technique, we eliminate an extra $\log n$ factor. This is at a cost of transforming root-log-diameters (c.f. Theorem \ref{thm:aware_fk_sample_complexity}) into linear diameter dependence. Retaining the \(\log n\) allows reducing this to root-\(\log\)-diameters.
\end{remark}

For the case of $f_k$-divergence, we can get a similar bound for the estimation error as follows.
For notation simplicity, define $k':= k/(k-1)$, $c_k(\delta):= (1 +k(k-1)\delta)^{1/k},$
and $a\vee b = \max\set{a,b}$.
\begin{theorem}[$f_k$-divergence constrained CAA adversary]\label{thm:aware_fk_sample_complexity}
Suppose Assumption
 \ref{assump:aware_lip_val_bounded_sps} is in force. Then, with $\cP = \set{\mu\ll\mu_0: D_{f_k}(\mu\|\mu_0)\leq \delta}$, when $n\geq 3\vee k$
$$\norm{\hat u - u^*} \leq 30\beta^2 n^{-\frac{1}{k'\vee 2}}c_k(\delta)^2\crbk{\frac{c_k(\delta)}{c_k(\delta)-1}\vee 2}\crbk{\frac{1}{k} + \sqrt{D+\log\frac{1}{\eta} + 2(d_\X+ d_\A)\log n}} $$
w.p. at least $1-\eta$. Here, $D := d_{\X}\log \crbk{1+3L\diam(\X)}+{d_{\A}}\log\crbk{1+3L\diam(\A)}$. 
\end{theorem} 

Perhaps surprisingly, Theorems \ref{thm:aware_Wd_sample_complexity} and \ref{thm:aware_fk_sample_complexity} demonstrate that, even though the goal is to learn the value function uniformly over a continuum of states and actions, and despite the additional adversarial robustness imposed by Wasserstein or $f_k$-divergence ambiguity sets, the estimation error $\|\hat{u} - u^*\|$ still decays at parametric rates. In particular, they do not suffer from the curse of dimensionality (i.e., rates of the form $n^{-c/|d_\X|}$), a phenomenon that typically arises when estimating infinite-dimensional objects. These results implies that, even in infinite state-action spaces and under robustness requirements, policy learning within the CAA setting of our DRSC framework remains statistically efficient.

\subsection{The Current-Action-Unaware Case}\label{section:UB_unaware}
For the CAU adversary case, we will operate under the following Assumption. 
\begin{assumption}\label{assump:unaware_lip_val_bounded_sps}
    Assume the following conditions: 
    \begin{enumerate}
        \item  $\X\subset\R^{d_{\X}},\W\subset\R^{d_{\W}}$ are equipped with the Euclidean distance with $\diam(\X)<\infty$.
        \item The action space $\A$ is finite, equipped with the 0-1 distance; i.e. $d(a,a') = \1\set{a \neq a'}$. 
        \item Assumption \ref{assump:cond_for_DPP} hold with $r_\vee = 1$.
        \item The mapping $(x,a)\ra r(x,a,\cd) + \bar u^*(f(x,a,\cd))$ is uniform $L$-Lipschitz
        \[\begin{aligned}
        &|r(x,a,w)- r(x',a',w)| + |\bar u^*(f(x,a,w)) - \bar u^*(f(x',a',w))| \leq L(|x-x'| +\1\set{a \neq a'}).\end{aligned}
        \]
    \end{enumerate}
\end{assumption}

As in the CAA setting, boundedness and Lipschitz continuity are natural assumptions that enable efficient uniform learning across the state space and control excessive oscillations in the value function. In the CAU case, however, we additionally assume that the action space $\A$ is finite. We will elaborate on this modeling choice in Remark \ref{rmk:finite_A}. Building on Assumption \ref{assump:unaware_lip_val_bounded_sps}, we are now prepared to establish convergence guarantees for the CAU adversary.

In the following Theorems \ref{thm:unaware_Wd_sample_complexity} and \ref{thm:unaware_fk_sample_complexity}, $\bar u^*$ and $\hat u$ are solutions to \eqref{eqn:def_action_unaware_DR_bellman_eqn} with corresponding Wasserstein and $f_k$ adversarial ambiguity sets $\cP$ and $\widehat\cP$ centered at $\mu_0$ and $\hat\mu$, respectively. 
\begin{theorem}[Wasserstein distance constrained CAU adversary]\label{thm:unaware_Wd_sample_complexity} Suppose Assumption
 \ref{assump:unaware_lip_val_bounded_sps} is in force. Also, assume that the cost is $c_\vee$-bounded. Then, with $\cP = \set{\mu:W_{c}(\mu,\mu_0)\leq\delta}$,
$$\norm{\hat u-\bar u^*}\leq c\crbk{ \beta \overline D_\vee\sqrt{d_{\X}+|\A|+1} +\beta^{3/2}\sqrt{\log{\frac{1}{\eta}}}}n^{-1/2}$$
w.p. at least $1-\eta$. Here $c = 32\sqrt{\pi}$ and $$\overline{D}_\vee:=   L\diam(\X) + 2\beta + c_\vee\delta\inv \beta +1.$$
\end{theorem}
\begin{theorem}[$f_k$-divergence constrained CAU adversary]\label{thm:unaware_fk_sample_complexity}

Suppose Assumption \ref{assump:unaware_lip_val_bounded_sps} is in force. Then, with $\cP = \set{\mu\ll\mu_0: D_{f_k}(\mu\|\mu_0)\leq \delta}$, when $n\geq 3\vee k$
$$\begin{aligned}
\norm{\hat u - \bar u^*}
&\leq 30\beta^2 n^{-\frac{1}{k'\vee 2}}c_k(\delta)^2\crbk{\frac{c_k(\delta)}{c_k(\delta)-1}\vee 2}\\
&\times \crbk{\frac{1}{k} + \sqrt{\overline D+\log\frac{1}{\eta} + 2(d_\X+ |\A|)\log n}} 
\end{aligned}$$
w.p. at least $1-\eta$. Here, the parameter $$\overline D:=d_\X\log \crbk{1+3L\diam(\X)} + |\A|\log\crbk{1+6\beta}.$$ 

\end{theorem}

\begin{remark}\label{rmk:finite_A}
As we discuss in Appendix \ref{a_sec:proof:thm:unaware_Wd_sample_complexity}, our proof can be extended to continuum action spaces under additional covering number requirements, yielding $n^{-1/2}$ convergence rates in those settings. In this paper, however, for CAU adversaries, we focus on the case $|\A|<\infty$ to obtain explicit dependencies on the dimensions, diameters, and the size of the action space. It is not clear whether the same parametric rate can be achieved when $\A \subset \R^d$ is compact and $\cQ = \cP(\cA)$. In such cases, an analysis analogous to the CAA setting yields an upper bound of $O(n^{-1/(2 \vee d_{\A})})$, but no matching lower bound is available. The exact rate in this setting remains an open question.  

Fortunately, if the controller’s set $\cQ$ is finite-dimensional (e.g., parametric, $\cQ = \{q_\theta \in \cP(\cA) : \theta \in \Theta \subset \R^d\}$, as is common when applying policy gradient methods with neural-network-parametrized policies), then even with continuous actions the same parametric convergence rate upper bound continues to hold.

\end{remark}

\section{Learning Optimal DRSC: Minimax Risks Lower Bounds}\label{sec:LB}
In this section, we study lower bounds on the finite sample minimax risk associated with learning the optimal value functions for both CAA and CAU adversary models. We demonstrate lower bounds that match the corresponding convergence rate upper bounds specified in the previous section.
\par Before we establish our lower bounds, we first introduce the finite sample minimax risk considered by this paper. For given value operator $\cK:\cP(\cW)\ra C(\X)$ and a class of probability measures $\cU\subset\cP(\cW)$, we define the $n$-sample minimax risk over $\cU$ of uniformly learning $\cK(\mu)$ as 
\[\begin{aligned}
&\mathfrak{M}_n(\cU,\cK) := \inf_{K}\sup_{\mu\in\cU}E_{\mu^n}\sup_{x\in\X}\abs{K(W_1,\ds,W_n)(x) - \cK(\mu)(x)}
\end{aligned}
\]
where $\mu^n = \mu\times \ds \times\mu$ is the $n$-fold product measure, and the first infimum is taken over all measurable functions $K: \W^n \ra C(\X)$.

As we will discuss in detail later, the operator \(\cK\) maps the center \(\mu\) of the Wasserstein distance and \(f_k\)-divergence constrained adversarial ambiguity sets to the solution of the Bellman equations \eqref{eqn:def_action_aware_DR_bellman_eqn} and \eqref{eqn:def_action_unaware_DR_bellman_eqn}. According to the dynamic programming principles, this solution corresponds to the optimal DRSC value. Therefore, \(\mathfrak{M}_n(\cU,\cK)\) represents the error incurred by the optimal learning algorithm (in terms of uniform performance over all centers of the ambiguity sets in \(\cU\)) for the optimal value when a data set with sample size \(n\) is available.

Specifically, we establish a lower bound on $\mathfrak{M}_n(\cU,\cK)$ for a simple value operator $\cK$ and a simple class of probability models $\cU$. We construct an instance with infinitely smooth reward and transition functions, where the minimax risk of learning the optimal robust value function achieves lower bounds that match the convergence rate upper bounds in Theorems~1–4. It turns out that to obtain such a matching lower bound, it suffices to let $\cU$ be the family of Bernoulli distributions with parameter $p \in [0,1]$.  

\par Concretely, for fixed dynamics $f$, reward $r$, and controller and adversarial ambiguity sets $\cQ$ and $\cP$, the value operator in the CAA (resp.\ CAU) case is defined by $\cK(\mu) = u^*$ (resp.\ $\cK(\mu) = \bar u^*$), where $u^*$ (resp.\ $\bar u^*$) is the solution to \eqref{eqn:def_action_aware_DR_bellman_eqn} (resp.\ \eqref{eqn:def_action_unaware_DR_bellman_eqn}) with $\mu_0$ replaced by $\mu$. With this definition, we consider the following DRSC instance.  

\begin{lemma}\label{lemma:LB_instance}
    Let $\X = \A = \W = [-1,1]$, $f(x,a,w) = w$, $r(x,a,w) = x$, and $\cQ = \cP(\cA)$. Then the solutions to \eqref{eqn:def_action_aware_DR_bellman_eqn} and \eqref{eqn:def_action_unaware_DR_bellman_eqn} are
    \begin{equation}\label{eqn:LB_instance_u}
        u^*(x) = \bar u^*(x) = x + \beta \inf_{\psi \in \cP} \int_{\W} w \,\psi(dw). 
    \end{equation}  
\end{lemma}

Using this instance, we can derive the following lower bounds on the minimax statistical risks in both the Wasserstein and the $f_k$ setting. 
\begin{theorem}[Lower bound under $W_2$-distance]\label{thm:Wd_lb} Let $\cP = \set{\mu:W_c(\mu,\mu_0)\leq \delta}$ with $c(x,y) = |x-y|^2$. The minimax risk of learning $u^*$ or $\bar u^*$ over any $$\cU \supset \set{\mu = p\delta_{\set{1}} + (1-p)\delta_{\set{0}}:p\in[0,1]}$$ is lower bounded by
$$\mathfrak{M}_n(\cU,\cK)\geq \frac{\beta }{32}n^{-\frac{1}{2}}. $$ 
\end{theorem}
\begin{remark}
We state the theorem only in terms of the $W_2$ distance. This is just for the convenience of calculations. It is not hard to see from the proof that using a Taylor expansion argument, for $W_p$ distances with $c(x,y) = |x-y|^p$, we have the same $n^{-1/2}$ rate, matching that in Theorem \ref{thm:aware_Wd_sample_complexity} and \ref{thm:unaware_Wd_sample_complexity}. 
Also, upon investigating the proof, one will find that the minimax risk of estimating the value function at \textbf{one single} $x$ has the same lower bound on the rate. 
\end{remark}

\begin{theorem}[Lower bound under $f_k$-divergence]\label{thm:fk_lb} 
Let $\cP = \set{\mu\ll\mu_0: D_{f_k}(\mu\|\mu_0)\leq \delta}$. Then there exist constants $C_1(k,\delta)$ and $C_2(k,\delta)$ that only depend on $k,\delta$ s.t. whenever $n\geq C_1(k,\delta)$, the minimax risk of learning $u^*$ or $\bar u^*$ over any $$\cU \supset \set{\mu = p\delta_{\set{1}} + (1-p)\delta_{\set{0}}:p\in[0,1]}$$ is lower bounded by
$$\mathfrak{M}_n(\cU,\cK) \geq  C_2(k,\delta)\beta n^{-\frac{1}{k'\vee 2}}.$$
\end{theorem}

\begin{remark}
The constants $C_1(k,\delta)$ and $C_2(k,\delta)$ are given in the proof in Appendix \ref{a_sec:proof:thm:fk_lb}.  This matches the convergence rate up to a $\sqrt{\log n}$ in Theorem \ref{thm:aware_fk_sample_complexity} and \ref{thm:unaware_fk_sample_complexity}. Again, estimating the value function at one single $x$ has the same lower bound. 
\end{remark}

Therefore, we have shown that even for the simple DRSC instance in Lemma \ref{lemma:LB_instance}, the minimax risk admits lower bounds that match the convergence rate upper bounds in Theorems~1–4. This establishes the tightness of our results and verifies the rates reported in Table~\ref{tab:summary_of_results}.

\section{Algorithm Design}
\label{sec:algoDesign}

Recall from our theoretical analysis that the solution $\hat u$ of the empirical Bellman equations achieves minimax-optimal convergence rates in estimating the true robust optimal values $u^\ast$ and $\bar u$. Motivated by this result, our algorithms are designed to approximate the empirical Bellman equations, $\bd{T}(u) = u$ and $\overline{\bd T}(u) = u$. In what follows, we focus on the $f_k$-divergence ambiguity setting and develop actor-critic algorithms with neural-network function approximation for DRSC under both CAA and CAU adversary settings.  

A key advantage of our approach is its seamless integration into standard RL pipelines: robustness can be introduced into existing actor-critic frameworks with only modest adjustments, without the need for extensive redevelopment. The method retains the familiar modular structure of separate neural policy and value function approximators, and can be implemented with only minor changes to widely used architectures and training loops. As a result, it is readily compatible with popular RL frameworks and easily adaptable across a wide range of domains.

\subsection{The CAA Case}

In the CAA setting, we parameterize the value function $u_\theta:\X \rightarrow \R$ and the policy $\pi_\eta:\X\rightarrow \A$ as neural networks. Leveraging the actor-critic framework, we alternate between Bellman error minimization and policy improvement steps, illustrated by Algorithm \ref{alg:caa}. Specifically, for a given policy-value pair $(\pi_\eta,u_\theta)$, the empirical Bellman operator is defined as:
\begin{align*}
&\bd T_{\eta,\theta}(x):=  \inf_{\psi\in \widehat\cP} \int_{\W} r(x,\pi_\eta(x),w) + \alpha u_\theta(f(x,\pi_\eta(x),w))\psi(dw),
\end{align*}
where the minimization is performed over the empirical $f_k$-divergence ambiguity set $\widehat{\cP}$ centered around the empirical measure $\hat \mu$, as described in previous sections.

\textbf{Strong Duality and Optimal Lagrange Multiplier:} To compute gradients of the empirical Bellman operator, we invoke strong duality \citep{duchi2021learning}, transforming the minimization over probability measures into an equivalent one-dimensional convex problem:
\begin{align*}
&\bd T_{\eta,\theta}(x)=  \sup_{\lambda\in\R}\bigg\{\lambda - c_k(\delta)\left[\int_{\W}(r(x,\pi_\eta (x),w)+ \alpha u_\theta(f(x,\pi_\eta(x),w))-\lambda)_+^{k'}\hat\mu(dw)\right]^{1/k'}\bigg\}.
\end{align*}
Here, $\lambda$ is the dual multiplier. Under mild assumptions, the maximization in $\lambda$ has a unique optimal solution $\lambda^*(\eta,\theta,x)$, which can be efficiently computed using bisection search. We define the intermediate operator for convenience:
\begin{align*}
&\widehat{\bd T}_{\eta,\theta}(\lambda,x)= \lambda - c_k(\delta)\left[\int_{\W} (r(x,\pi_\eta(x),w) + \alpha u_\theta(f(x,\pi_\eta(x),w))-\lambda)_+^{k'}\hat\mu(dw)\right]^{1/k'}.
\end{align*}

\textbf{Bellman Error Minimization:} Given a fixed policy $\pi_\eta$, we minimize the squared $L^2$ Bellman error:
\[
\min_{\theta} \int_{\X} [u_\theta(x)-\bd T_{\eta,\theta}(x)]^2 \nu(dx),
\]
where $\nu$ is a user-specified probability measure supported on the entire state space $\X$. To compute gradients, we utilize the envelope theorem to get that:
\begin{equation}\label{eqn:op_grad_envelope}
\nabla_\theta \bd T_{\eta,\theta}(x)=\nabla_\theta\widehat{\bd T}_{\eta,\theta}(\lambda^*(\eta,\theta,x),x).
\end{equation}

\par To minimize Bellman Error we update the $\theta$ using first-order algorithms. For illustrative purposes, we formulate the mini-batch stochastic gradient descent in this context. At each iteration, we independently sample states $X_i\sim\nu$, compute the optimal multipliers $\lambda_i^*$ in parallel, and update the parameters as follows:
\begin{equation}\label{eqn:bem_update}
    \theta_{t+1} = \theta_t - \ell_t \frac{1}{n}\sum_{i=1}^{n} \nabla_\theta[(u_\theta(X_i) - \bd T_{\eta,\theta}(x))^2],
\end{equation}
where the gradient w.r.t. $\theta$ is computed using the chain-rule and \eqref{eqn:op_grad_envelope}. 

\textbf{Policy Improvement: }
Given a fixed value function $u_\theta$, the policy $\pi_\eta$ is improved by solving the following optimization problem with first-order methods:
\[
\max_\eta \int_\X \bd T_{\eta,\theta}(x) \nu(dx)
\]
with gradients computed similarly using the envelope theorem. Again, we update parameters via stochastic gradient ascent:
\begin{equation}\label{eqn:pi_update}
    \eta_{t+1} = \eta_t + \ell'_t \frac{1}{n}\sum_{i=1}^{n}\nabla_\eta\widehat{\bd T}_{\eta,\theta}(\lambda_i^*,X_i). 
\end{equation}

\begin{algorithm}[ht]
\caption{Robust Actor–Critic Algorithm for CAA}\label{alg:caa}
\begin{algorithmic}[1]
\Require Step–size schedules $\{\ell_t\}_{t\ge0},\{\ell'_t\}_{t\ge0}$, batch size $n$, initial parameters
         $(\theta_0,\eta_0)$, distribution $\nu$ on $\X$
\vspace{0.3em}
\Statex \textbf{Repeat until convergence:}
\Statex \rule{\linewidth}{0.4pt}
\Statex \textbf{1. Bellman error minimization (critic) update}
\State Sample states $\{X_i\}_{i=1}^{n}\overset{\text{i.i.d.}}{\sim}\nu$
\For{$i=1,\dots,n$ \textbf{in parallel}}
    \State $\displaystyle\lambda_i^{\!*}\;\leftarrow\;
        \arg\max_{\lambda\in\R}\;
        \widehat{\bd T}_{\eta_t,\theta_t}(\lambda,X_i)$ \Comment{bisection search}
\EndFor
\State Update $\theta$ as in Equation \eqref{eqn:bem_update}
\vspace{0.4em}

\Statex \textbf{2. Policy improvement (actor) update}
\State Resample (or reuse) $\{X_i\}_{i=1}^{n}\sim\nu$
\For{$i=1,\dots,n$ \textbf{in parallel}}
    \State Recompute $\displaystyle
        \lambda_i^{\!*}\leftarrow
        \arg\max_{\lambda}\widehat{\bd T}_{\eta_t,\theta_{t+1}}(\lambda,X_i)$
\EndFor
\State Update $\eta$ as in Equation \eqref{eqn:pi_update}
\State $t\;\leftarrow\;t+1$
\Statex \rule{\linewidth}{0.4pt}
\vspace{0.2em}
\Return final parameters $(\theta_{t},\eta_{t})$
\end{algorithmic}
\end{algorithm}

\subsection{The CAU Case with Continuous Action}

While our theoretical analysis focuses primarily on finite action models in the CAU setting, practical applications can involve continuous action spaces. Hence, we propose an extended algorithm suitable for continuous state-action problems. The CAU case is inherently more complex since deterministic policies are generally suboptimal, necessitating the consideration of randomized policies. To address this, we propose a novel \textit{generative} policy approach. Specifically, we view a policy as a generative model that produces randomized actions given a state $x\in\X$ by utilizing an external source of randomness. 

\par Concretely, we define the policy as a mapping $\pi_\eta: \R^d \times\X\rightarrow \A$, where an action is generated according to $\pi_\eta(N,x)$ and $N\sim N(0,I)$ is a standard normal random vector independent of the state. For each state $x$, we sample a normal vector $N$ of $d$ dimensions. Due to space constraints, a detailed discussion of the algorithm for the CAU case is provided in Appendix \ref{a_sec:CAU_alg}.

\section{Experiments}
\label{sec:experiment}

We provide two sets of experiments using the DRSC framework with CAA and CAU adversaries proposed above. The first one is a study on inventory control (Example \ref{exmpl:inventory} in Section \ref{subsec:modeling_examples})  with real-world data, and the second one is a simulation study on portfolio optimization (Example \ref{exmpl:portfolio} in Section \ref{subsec:modeling_examples}). For each setting, we solve the robust value function and policy using the algorithm described in Section \ref{sec:algoDesign}, and compare it with the non-robust optimal policy.

\subsection{Inventory Control}
Inventory management is a cornerstone of retail and supply chain operations. Managers must balance the cost of carrying excess stock against the risk of lost sales when demand outstrips supply. This trade-off becomes particularly challenging when customer demand is uncertain and subject to shifts caused by seasonality, promotions, or broader economic disruptions. Traditional models often assume that the demand process is i.i.d. across time, but in practice, this assumption rarely holds. Unexpected fluctuation, such as unaccounted shifts in consumer preferences or disruptions in the broader economy, can cause substantial performance degradation when managers rely solely on non-robust policies trained under the assumption of i.i.d. demand.

In this section, we empirically demonstrate how managers can benefit from our distributionally robust approach to improve decision-making in a data-driven context. By explicitly accounting for non-stationary shifts and potential adversary behavior in the demand process, our framework enables the design of effective and robust ordering policies that hedge against worst-case scenarios.

\subsubsection{Setting}

We consider Example \ref{exmpl:inventory} in Section \ref{subsec:modeling_examples}, where a decision maker must determine the quantity of a single product to order over discrete time periods $t = 0,1,2,\dots$. Let $X_t \in \R_{\ge 0}$ denote the inventory level at the beginning of period $t$, and let $A_t \in \R_{\ge 0}$ be the quantity ordered at the beginning of that period. We assume that orders placed at time $t$ arrive instantaneously and are available to satisfy demand in the same period.

\par During each period, a random demand $W_t \in \R_{\ge 0}$ is realized, and unmet demand is either backlogged or lost. For simplicity, we assumed the unmet demand is directly lost. 

The performance of a policy is measured via the expected infinite horizon discounted total cost. Note that our theorems can be extended to the case where the reward is the function of $X_t, A_t, W_t$ as shown before.

We use the open Rossmann Store Sales dataset from Kaggle \footnote{https://www.kaggle.com/competitions/rossmann-store-sales/}.
It contains the demand data from 2013 to 2016. For one store, we use the earliest 80\% data as the training set from which we calculate the empirical distribution of $W_t$; we test the trained robust policies on the time sequence of the latest 20\% data. To confirm the generalizability of our results, we repeat the same experiment procedure on 5 different stores and report all the results. We set $h=0.1, c=1, p=2, \alpha=0.9$. More implementation details can be found in Appendix \ref{a_sec:result_exp}.

\subsubsection{Results}
The results are shown in Table \ref{tb:inventory}. Note that in this case, the same-day demand does not react to $A_t$, and it shows a trend of auto-correlation, so CAU is appropriate. For each CAU policy, we report the variance of the return by repeatedly sampling the normal random sequence 50 times. In each case, we could calculate the explicit solution of the non-robust optimal policy \cite{scarf1960optimality}, and compare the performance of the CAA and CAU policy with it. We find that
\begin{enumerate}
    \item For properly small and mild $\delta$, both CAA and CAU robust policies outperform the non-robust optimal policy, which also serves as an illustration of the distribution shift in demand over time.
    \item For small $\delta$, CAA outperforms CAU in most cases.
    \item For mild and large $\delta$, CAU outperforms CAA, and it also has the hightest returns in 4 out of 5 cases.
\end{enumerate}

Note that the non-robust policy is a base stock policy (S-policy). And we find that the CAA robust policies are also S-policies, which corresponds to the explicit solutions in \cite{wagner2018robust}. We also see a positive relation between the size of the uncertainty set and randomization - the CAU optimal policy becomes more randomized as the uncertainty set get larger. However, the randomization is not always better, especially for small $\delta$, because the distribution shift may not be in the adversarial direction. The plots of CAA and CAU policies are in Appendix \ref{a_sec:result_exp}.




\begin{table}[htb!]
\centering
\caption{Return of CAA and CAU policies for 5 stores}

\begin{tabular}{| c | c | c | c | c | c |}
\hline
{\texttt{Return}~($\times 10^{3}$)} & $\delta=0.001$ & $\delta=0.01$ & $\delta=0.1$ & $\delta=0.5$ & $\delta=1$ \\\hline
\texttt{CAA} & 38.04 & 38.38 & \textbf{38.47} & 37.05 &  36.61\\\hline
\texttt{CAU} & $37.95\pm0.01$  & $38.26\pm0.02$ & \textbf{38.54}$\pm0.03$ & $37.55\pm0.1$ & $36.86\pm0.2$\\\hline
\texttt{$\pi_0^*$} & \multicolumn{5}{c|}{37.85} \\
\hline

\multicolumn{6}{c}{}\\
\hline
{\texttt{Return}~($\times 10^{3}$)} & $\delta=0.001$ & $\delta=0.01$ & $\delta=0.1$ & $\delta=0.5$ & $\delta=1$ \\
\hline
\texttt{CAA} & 81.72 & 81.83 & \textbf{81.97} & 76.48 & 70.65\\
\hline
\texttt{CAU} & $81.46\pm0.01$  & $81.79\pm0.02$ & \textbf{82.18}$\pm0.03$ & $77.76\pm0.03$ &$75.04\pm0.03$\\\hline
\texttt{$\pi_0^*$} & \multicolumn{5}{c|}{81.60} \\
\hline

\multicolumn{6}{c}{}\\
\hline
{\texttt{Return}~($\times 10^{3}$)} & $\delta=0.001$ & $\delta=0.01$ & $\delta=0.1$ & $\delta=0.5$ & $\delta=1$ \\
\hline
\texttt{CAA} & 31.75 & 32.21 & \textbf{33.17} & 29.67 & 29.64\\
\hline
\texttt{CAU} & $31.46\pm0.01$  & $31.94\pm0.02$ &\textbf{33.38}$\pm0.03$ & $31.26\pm0.2$ & $30.48\pm0.3$\\\hline
\texttt{$\pi_0^*$} & \multicolumn{5}{c|}{31.16} \\
\hline

\multicolumn{6}{c}{}\\
\hline
{\texttt{Return}~($\times 10^{3}$)} & $\delta=0.001$ & $\delta=0.01$ & $\delta=0.1$ & $\delta=0.5$ & $\delta=1$ \\
\hline
\texttt{CAA} & 62.28 & 63.12 & \textbf{64.47} & 62.02 & 54.70\\
\hline
\texttt{CAU} & $62.13\pm0.01$  & $62.95\pm0.02$ & \textbf{64.42}$\pm0.03$ & $62.97\pm0.1$ & $57.53\pm1.0$\\\hline
\texttt{$\pi_0^*$} & \multicolumn{5}{c|}{62.26} \\
\hline

\multicolumn{6}{c}{}\\
\hline
{\texttt{Return}~($\times 10^{3}$)} & $\delta=0.001$ & $\delta=0.01$ &  $\delta=0.1$ & $\delta=0.5$ & $\delta=1$ \\
\hline
\texttt{CAA} & 61.20 & 62.16 & \textbf{63.11} & 60.98 & 60.20\\
\hline
\texttt{CAU} & $61.93\pm0.01$  & $61.99\pm0.02$ & \textbf{63.17}$\pm0.02$ &$61.24\pm0.03$ & $60.74\pm0.1$\\\hline
\texttt{$\pi_0^*$} & \multicolumn{5}{|c|}{59.66} \\
\hline
\end{tabular}

\label{tb:inventory}
\end{table}

Our findings highlight several takeaways for managers:

\begin{enumerate}
    \item \textbf{Value of Robustness under Demand Shifts.} When demand patterns shift over time, robust policies consistently outperform the classical non-robust policy. This underscores the importance of preparing for distributional changes rather than optimizing exclusively for historical distributions.
    \item \textbf{Different Forms of Robustness.} For relatively small deviations from historical demand, CAA performs best; for larger deviations, CAU dominates. This suggests that managers should calibrate the level of robustness to the degree of environmental volatility they expect.
    \item \textbf{Trade-off Between Randomization and Performance.} We observe that CAU policies introduce more variability in ordering decisions as uncertainty grows. While this can be beneficial when facing severe disruptions, excessive randomization may reduce performance when demand is relatively stable. Hence, robustness should be deployed thoughtfully, balancing protection against downside risk with efficiency in normal times.
\end{enumerate}

\subsection{Portfolio Optimization}
In addition to the real data case, we want to test our proposed method on a more challenging case, where several constraints are violated. We find our method still applicable to this hard case with some numerical techniques.

\subsubsection{Motivation}
Portfolio allocation is one of the most fundamental decisions in finance, requiring investors to balance consumption today against investment for tomorrow. A classical result from the Merton model \citep{samuelson1975lifetime} shows how investors allocate between risk-free and risky assets under known return distributions. However, in practice, asset returns are uncertain, non-stationary, and subject to shocks such as recessions, market bubbles, or geopolitical events. Standard optimization methods that rely on historical averages may therefore expose investors to substantial downside risk.

Our framework addresses this challenge by designing policies that remain effective even when the return distribution shifts. Specifically, we test how distributionally robust policies—modeled under our CAA and CAU approaches—alter portfolio allocation strategies in the face of adversarial shocks. This provides insights into how investors should adjust their risk exposure when market conditions deviate from expectations.

\subsubsection{Setting}

We consider Example \ref{exmpl:portfolio} in Section \ref{subsec:modeling_examples}, where we want to manage a portfolio of $m$ assets where the time is divided into discrete time periods $t = 0,1,2,...$ (not necessarily uniformly spaced in real time). $X_t \in \R^m$ denotes the portfolio (or vector of positions) at time $t$, where $X_{t,i}$ is the dollar value of
asset $i$ at the beginning of time period $t$: $X_{t,i} > 0$ and $X_{t,i} < 0$ mean a long position and a short position in asset $i$, respectively. We can buy and sell assets at the beginning of each time period. Let $A_t \in \R^m, t = 0, 1, \dots$ be the resulting percentage of the $m$ assets after trading at $t$: $A_{t,i} > 1$ or $0 \le A_{t,i} < 1$ means buying or selling asset $i$ at the beginning of time period $t$, respectively. The performance of the policy is measured in terms of the expected infinite horizon discounted total utilities. 

We consider the 2-asset case, i.e. $m=2$. This is a common case where one asset is risk-free while the other is risky. We set the risk-free return to be 1.04, and the risky return distribution to be lognormal with expected return around 1.13; i.e.  $r_f = 1.04, ~\psi_0(r_{risky}) \sim log\mathcal{N}(0,0.5)$.

Assume the reward is the utility from consumption as in Merton model \cite{samuelson1975lifetime}, $r(X_t, A_t, W_t) = \frac{C_t^{\gamma}}{\gamma}$. We set $\gamma=0.5$, and the discount factor $\alpha=0.9$. Also, we impose another assumption that the total consumption should not exceed the total assets; $\bd{1}^\top X_t -\bd{1}^\top A_t \le \bd{1}^\top X_t$, i.e. $\bd{1}^\top A_t \ge 0$. And we assume that $\bd{1}^\top X_0 \ge 0$ in the beginning.

Note that based on our constraints, once $\bd{1}^\top X_t \le 0$, we must have $0 \le \bd{1}^\top A_t \le \bd{1}^\top X_t \le 0$. Thus, instead of waiting for $\bd{1}^\top X_t$ to become positive again (which may never happen), we end the process immediately.

\subsubsection{Results}

\begin{figure}[t!]
    \centering
    \begin{subfigure}[t]{0.5\textwidth}
        \centering
        \includegraphics[height=2.5in]{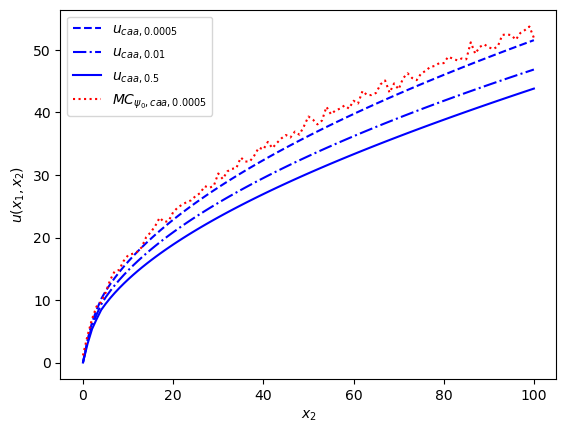}
        \caption{Slice of CAA value function at $x_1=0$}
    \end{subfigure}%
    ~ 
    \begin{subfigure}[t]{0.5\textwidth}
        \centering
        \includegraphics[height=2.5in]{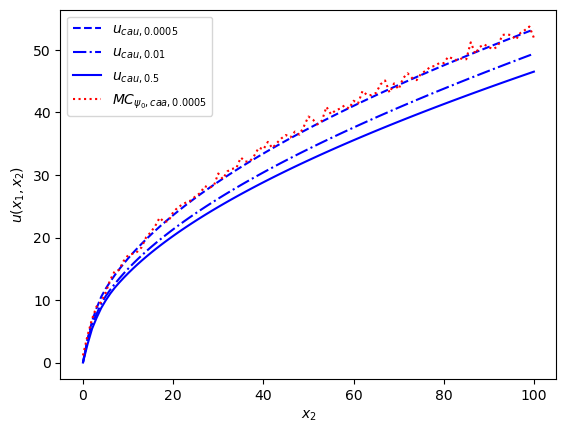}
        \caption{Slice of CAU value function at $x_1=0$}
    \end{subfigure}
    \caption{Robust value functions with various sizes of the uncertainty set.}
    \label{fig:port_value}
\end{figure}

\begin{figure}[t!]
    \centering
    \begin{subfigure}[t]{0.5\textwidth}
        \centering
        \includegraphics[height=2.5in]{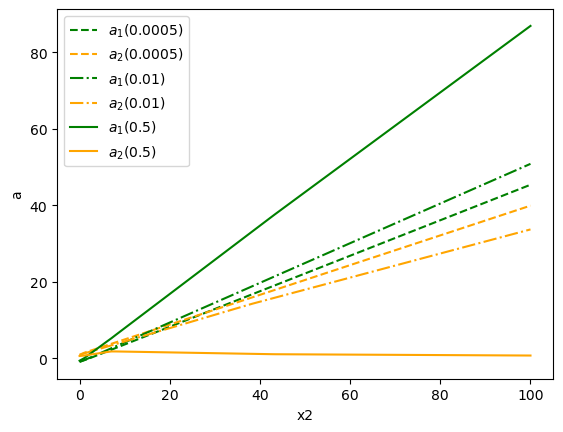}
        \caption{Slice of CAA policy at $x_1=0$}
    \end{subfigure}%
    ~ 
    \begin{subfigure}[t]{0.5\textwidth}
        \centering
        \includegraphics[height=2.5in]{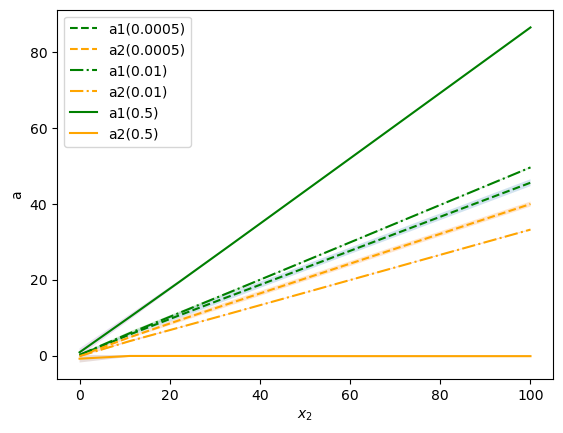}
        \caption{Slice of CAU policy at $x_1=0$}
    \end{subfigure}
    \caption{Robust policies with various sizes of the uncertainty set.}
    \label{fig:port_policy}
\end{figure}

\begin{figure}[t!]
    \centering
    \begin{subfigure}[t]{0.5\textwidth}
        \centering
        \includegraphics[height=2.5in]{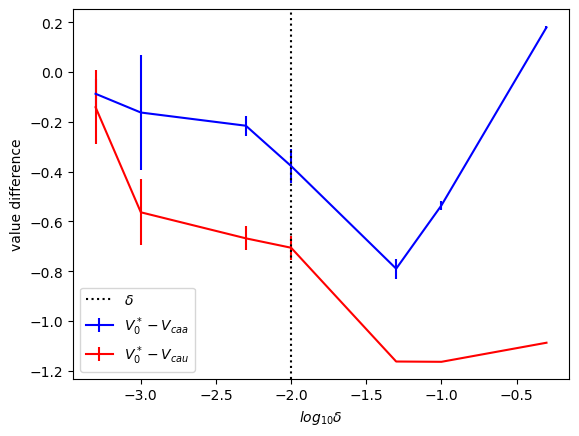}
    \end{subfigure}
    \caption{Value difference between non-robust optimal policy and two robust policies.}
    \label{fig:port_v_delta}
\end{figure}

We compare three strategies:  
\begin{itemize}
    \item the classical non-robust policy based on historical distributions,  
    \item the robust CAA policy, and  
    \item the robust CAU policy.  
\end{itemize}
Robust policies are trained under varying levels of uncertainty $\delta$, reflecting different degrees of market stress or volatility. The results are shown in Figure \ref{fig:port_value}, \ref{fig:port_policy} and \ref{fig:port_v_delta}. In Figure \ref{fig:port_value}, ``$\text{MC}_{\psi_0, \text{caa}, 0.0005}$'' is the Monte Carlo value estimation of the robust policy with $\delta=0.0005$ in the original environment $\psi_0$ (in order to verify the correctness of the converged value and policy).

We have the following observations:
\begin{enumerate}
    \item The robust value becomes smaller as the size of the uncertainty set $\delta$ increases. For the same $\delta$, the value of CAU policy is higher than that of CAA policy. This is because CAU allows randomized policies which is at least no worse than CAA policy.
    \item The robust policy becomes more conservative as the size of the uncertainty set $\delta$ increases, in the sense that more risk-free and less risky asset would be held.
    \item We set the real $\delta$ to be $10^{-2}$, and we consider the $\chi^2$ divergence. This means that the test distribution of the risky asset is $log\mathcal{N}(-\sqrt{\ln(1+2\delta)}/2,0.5)$. With the adversarial distribution shift, the CAU policy is consistently better than the CAA policy. With small and mild $\delta$, both robust policies are better than the non-robust policy.
\end{enumerate}

As mentioned before, because the trader here is small relative to the market and returns are driven by external factors, it is reasonable to use the CAU framework.

Our analysis yields several finance-relevant takeaways:

\begin{enumerate}
    \item \textbf{Flight to safety under heightened uncertainty.} As the size of the ambiguity set $\delta$ increases, both robust policies reallocate wealth away from the risky asset toward the risk-free asset. This mirrors the “flight to quality” observed in financial markets during periods of elevated volatility, where investors naturally hedge by holding safer assets. Our results formalize this behavior as the optimal response to distributional ambiguity.
    
    \item \textbf{Adaptive diversification through CAU policies.} The CAU framework, which allows randomized allocations across assets, consistently delivers higher utility than CAA when market shocks are severe. This corresponds to a form of adaptive diversification—spreading exposure in a probabilistic manner rather than committing deterministically to one allocation. In practice, this finding suggests that fund managers facing crisis conditions may benefit from dynamic or mixed strategies that preserve flexibility in the face of adversarial return scenarios.
    
    \item \textbf{Robustness improves risk-adjusted performance.} For small to moderate levels of distributional shifts, both CAA and CAU policies outperform the classical non-robust benchmark. In finance terms, this means that incorporating robustness can improve Sharpe-like efficiency: investors attain higher expected utility (adjusted for risk aversion) without incurring additional downside exposure. Robust optimization thus not only insures against tail events but also enhances performance in moderately perturbed environments.
\end{enumerate}

Together, these insights demonstrate that robust portfolio policies provide a disciplined way to operationalize risk management principles widely practiced in finance—hedging in uncertain markets, diversifying under stress, and improving risk-adjusted returns when market conditions deviate modestly from historical patterns.

\subsection*{Acknowledgments}
    The material in this paper is partly supported by the Air Force Office of Scientific Research under award number FA9550-20-1-0397. Support from NSF 2229012, 2312204, 2312205, 2403007, 2419564, ONR 13983111, 13983263, and 2025 New York University Center for Global Economy and Business grant is also gratefully acknowledged.
\bibliographystyle{apalike}
\bibliography{bibs/DR_MDP,bibs/proofs_ref,bibs/rl,bibs/drrl,bibs/applications,bibs/dro}

\newpage
\appendix

\section{Formulations of Distributionally Robust Stochastic Control}\label{a_sec:DRSC_formulation}
\par Let $\X,\A,\W$ be Polish spaces and $(\X,\cX),(\A,\cA),(\W,\cW)$ equip them with the Borel $\sigma$-fields. Let $\cP(\cA)$ and $\cP(\cW)$ be the set of probability measures on $(\A,\cA)$ and $(\W,\cW)$, respectively. Endow them with the topology of weak convergence; i.e. $\mu_n\Ra\mu$ if $\int fd\mu_n\ra \int fd\mu $ for all bounded continuous $f$. Then, $\cP(\cA)$ and $\cP(\cW)$ are separable, as $(\A,\cA)$ and $(\W,\cW)$ are separable. 
\par We now present our distributionally robust stochastic control formulation. Let $\Omega = \X\times (\A\times \W)^{\Z_+}$ and $\cF$ is the $\sigma$-field generated by cylinder sets. A canonical element $\omega\in\Omega$ is $\omega = (x_0,a_0,w_0,a_1,w_1\ds a_t,w_t\ds)$, $x_0\in\X$, $w_k \in \W$, and $a_k\in\A$, $\forall k\geq 0$. 
\par Let $W:= \set{W_t:{t\geq 0}}$ and $A:= \set{A_t:{t\geq 0}}$ be the processes of point evaluation of $\set{w_t:t\geq 0}$ and  $\set{a_t:t\geq 0}$, respectively; i.e. 
$$W_t(\omega) = w_t,\quad A_t(\omega) = a_t.$$ 
Finally, define the process $X:=\set{X_t:t\geq 0}$ by the stochastic recursion $X_0(\omega) = x_0$ and for each $t\geq 0$
$$X_{t+1} = f(X_{t},A_t,W_t).$$
\par We refer to $X$ as the controlled state process, $A$ as the action process, and $W$ as the exogenous noise process. In the classical stochastic control setting, a typical assumption is that the noise process $W$ consists of i.i.d. $W_t$ under any probability measure of interest on $(\Omega,\cF)$. In our setting, however, the adversary can dynamically perturb the distribution of $W_t$ based on some or all historical information, potentially making it a general stochastic process with arbitrary dependent structure. 
\subsection{Admissible Policies}
In this section, we rigorously formulate the controller and adversarial policies under the DR Stochastic control framework. We formulate the the controller and the adversary policies so that, collectively, they will give rise to a unique probability measure on $(\Omega,\cF)$. At a high level, for each and every $t\geq 0$, the controller and the adversary choose the conditional distributions of $A_t$ and $W_t$ respectively, given their available information. 
\par Let us define the following notations. For measures $\mu,\nu$ on $(\C,\cC)$, we write $\mu(dc) = \nu(dc)$ if $\mu(C) = \nu(C)$ for all $C\in\cC$.

\par For $t\geq 0$, define controller's history $$\bd{H}_t := \set{h_t = (x_0,a_0,\ds,a_{t-1},x_t):x_k\in \X,a_k\in \A, \forall k}. $$ 
and the adversarial history $$\bd{G}_t := \set{g_t = (x_0,a_0,\ds,x_t,a_t):x_k\in \X,a_k\in \A,\forall k }.$$
For convenience, we let $\bd{H}_{-1}= \bd{G}_{-1} = \varnothing$. 
\par Define the history random elements 
$$H_t(\omega) := h_t = (x_0,a_0,\ds,x_t)\in\bd{H}_t\quad \text{and}\quad G_t(\omega) := g_t = (x_0,a_0,\ds,x_t,a_t)\in\bd{G}_t$$ where $x_{k+1} = f(x_{k},a_k,w_k)$ for $k = 1,\ds,t-1$, recursively.
\subsubsection*{Admissible Controller's Policies}
\par A \textit{decision} of the controller $\pi_t$ at time $t$ is a (product space) Borel measurable function $\pi_t:\bd{H}_t\ra\cP(\cA)$. This is seen as the conditional distribution of $A_t$ given history $H_t=h_t$, hence we write $\pi_t(da|h_t)$. A \textit{policy} of the controller $\pi = (\pi_0,\pi_1,\ds )$ is a sequence of decisions. The largest possible policy class under this framework is the the history-dependent unconstrained controller's policy class:
$$\PiH := \set{\pi = (\pi_0,\pi_1,\ds):\pi_t\in m\set{\bd{H}_t\ra\cP(\cA)}}$$
where $m\set{\bd{H}_t\ra\cP(\cA)}$ denote the set of Borel measureable functions. 
\par To increase the modeling flexibility of our DR stochastic control framework, we consider constraints on the controller in terms of information availability and admissible set of controller's decisions.
\par We say that a controller's policy $\pi = (\pi_0,\pi_1,\ds)$ is Markov if for each and every $t\geq 0$, 
$$\pi_t(da|g_{t-1},x_t) = \pi_t(da|g'_{t-1},x_t)$$
for any $g_{t-1},g_{t-1}'\in\bd{G}_{t-1}$ and $x_t\in\X$; i.e. given $x_t$, the distribution of the action is independent of the history $g_{t-1}$. Therefore, through an abuse of notation, we can write $\pi_t(da|x)$ when the decision is Markov. Denote the set of Markov controller's policies by $\PiM$. 

\par Moreover, $\pi$ is said to be time-homogeneous (or stationary Markov) if $$\pi_t(da|g_{t-1},x) = \pi_s(da|g_{s-1}',x)$$ for every $s,t\geq 0$ and $g_{t-1}\in\bd{G}_{t-1}$, $g_{s-1}'\in\bd{G}_{s-1}$ and $x\in\X$; i.e. $\pi$ is Markov and invariant in time. As in the Markov case, we can write $\pi(da|x)$ when the decision is time-homogeneous. We denote the set of time-homogeneous controller's policies by $\PiS$. 
\par We further allow the controller to be constrain to choose its decision $\pi_t(da|h_t)\in\cQ$ from a admissible subset $\cQ\subset\cP(\cA)$ that is Borel measureable. This can be done under any information structures defined above. We denote such constrained controller with the corresponding information availability as $\Pi_{\mrm{U}}(\cQ)$ where $\mrm{U} = \mrm{H},\mrm{M},\mrm{S}$. 

\subsubsection*{Admissible Adversarial Policies}

A decision of the adversary $\gamma_t$ at time $t$ is a measurable function $\gamma_t:\bd{G}_t\ra\cP(\cW)$, where we write $\gamma_t(dw|g_t)$ and note that it signifies the conditional distribution of $W_t$ given $G_t = g_t$. An adversarial policy $\gamma = (\gamma_0,\gamma_1,\ds )$ is a sequence of adversarial decisions. This forms the the history-dependent unconstrained adversary's policy class:
$$\GammaH := \set{\gamma = (\gamma_0,\gamma_1,\ds):\gamma_t\in m\set{\bd{G}_t\ra\cP(\cW)}}.$$
\par We define an adversary's policy to be Markov if $\gamma_t(da|g_{t-1},x_t,a_t) = \gamma_t(da|g'_{t-1},x_t,a_t)$ for any $g_{t-1},g_{t-1}'\in\bd{G}_{t-1}$ and $x_t\in\X ,a_t\in \A$. Further, an adversary's policy is time-homogeneous (or stationary Markov) if $\gamma_t(da|g_{t-1},x,a) = \gamma_s(da|g_{s-1}',x,a)$ for every $s,t\geq 0$ and $g_{t-1}\in\bd{G}_{t-1}$, $g_{s-1}'\in\bd{G}_{s-1}$ and $x\in\X,a\in \A$. As in the controller setting, we write $\gamma_t(dw|x,a)$ and  $\gamma(dw|x,a)$ for Markov and time-homogeneous adversary decisions, respectively. Denote the Markov and time-homogeneous adversarial policy classes by $\GammaM$ and $\GammaS$, respectively. 
\par As for the controller's case, we allow the adversary to be constrain to choose $\gamma_t(da|g_t)\in\cP$ from a admissible subset $\cP\subset\cP(\cW)$ that is Borel measureable. We denote such constrained adversarial policy classes with the corresponding information availability as $\Gamma_{\mrm{U}}(\cP)$ where $\mrm{U} = \mrm{H},\mrm{M},\mrm{S}$.

\begin{remark}
Notice that in this model, the history-dependent controller cannot directly use the realized action $w_t$ of the adversary to make its decision. This should be compared to the stochastic game settings \cite{gonzalez2002minimax} in which either player observes the action of the other and makes a decision based on such observation. However, this model can include the settings for which both players see the action of each other by considering a new state process $z_t = (x_t,w_{t-1})$ and defining the state space and histories using $z_t$ instead of $x_t$. 

\par We also note that in general settings for which the modeler decides to construct $f$ so that $W_{k-1}\notin \sigma(X_{k})$ for each $k\leq t$. Then, this is as if the adversary cannot use its  historical actions $\set{W_k:k\leq t-1}$ to decide the distribution of the current action $W_t$. 

\end{remark}

\subsubsection*{Current-Action-Unaware Adversary}
Consider adversarial policy $\gamma = (\gamma_0,\gamma_1,\ds)\in\Gamma_{\mrm{H}}$. Because in general, the distribution of $W_t$ depends on the current action $a_t$ through $g_t$, i.e. $W_t\sim \gamma_t(dw|g_t)$, we say that they are \textit{current-action-aware} (CAA). However, in many settings, the adversary cannot base its decision on the current action $a_t$. Such adversary is characterized by the following concept of \textit{current-action-unaware} (CAU) decisions. 

\par We say that a adversary's decision $\gamma_t$ is current-action-unaware if
\begin{equation}\label{eqn:curr_action_unaware_adv_decision}
\gamma_t(dw|g_t) = \gamma_t(dw|h_t,a')    
\end{equation} for all $a'\in\A$, where $g_t = (h_t,a_t)$. Then the set of history dependent current-action-unaware adversary with constraint set $\cP$ is a subset $\barGammaH\subset\Gamma_{\mrm{H}}$ defined by $$\barGammaH(\cP):= \set{\gamma = (\gamma_0,\gamma_1,\ds): \gamma_t\in m\set{\bd{G}_t\ra \cP}, \gamma_t(dw|g_t) = \gamma_t(dw|h_t,a'), \forall a'\in A}$$
When $\gamma$ is independent of the current action, we write $\bar \gamma_t(dw|h_t):= \gamma_t(dw|h_t,a)$. Hence, we have $\bar\gamma = (\bar\gamma_0,\bar\gamma_1,\ds)\in\barGammaH(\cP)$. 
\par This can be easily generalized to the Markov and time-homogeneous settings by $\barGammaM(\cP):= \barGammaH(\cP)\cap \GammaM(\cP)$ and $\barGammaS(\cP):= \barGammaH(\cP)\cap \GammaS(\cP)$, consisting of Markov and time-homogeneous policies for which the decision at any time is current-action-unaware as defined in \eqref{eqn:curr_action_unaware_adv_decision}.

\subsection{The Distributionally Robust Stochastic Control Problem}

\par Given an initial distribution $\mu_0$ on $\cP(\cX)$ and a pair of controller's and adversary's policy $(\pi,\gamma)$, define a probability measure $P_{\mu}^{\pi,\gamma}$ on $\Omega$ as follows. For cylinder sets of the form 
$$
C_t:= B_0\times Y_0\times \ds\times B_t\times Y_t\times \A\times \W \times \A\times \W\ds
$$
for some $B_k\in \cA$ and $Y_k\in\cW$ for each $k\leq t$, define
\begin{equation}\label{eqn:def_prob_meas_from_pi_gamma}
P_{\mu}^{\pi,\gamma}(C_t):= \int_{\X}\int_{B_0}  \int_{Y_0}\int_{B_1} \ds \int_{Y_t}\gamma_t(dw_t|g_t) \ds  \pi_1(da_1|h_1)\gamma_0(dw_0|g_0) \pi_0(da_0|h_0)\mu_0(dx_0).
\end{equation}
This uniquely extends to a probability measure on $(\Omega,\cF)$. Let $E_{\mu}^{\pi,\gamma}$ denote the expectation under $P_{\mu}^{\pi,\gamma}$. 
\par The distributionally robust stochastic control (DRSC) paradigm under this formulation where an adversary perturbs the exogenous driving randomness aims to find the infinite horizon discounted maxmin value function
\begin{equation}\label{eqn:DRSC_value_func}
v^*(\mu,\Pi,\Gamma):= \sup_{\pi\in\Pi} \inf_{\gamma\in \Gamma} v(\mu,\pi,\gamma), \quad v(\mu,\pi,\gamma):=E_{\mu}^{\pi,\gamma} \sum_{t=1}^\infty\alpha^tr(X_t,A_t,W_t)
\end{equation}
subject to $X_{k+1} = f(X_k,A_k,W_k)$, $\forall k$. Here, the admissible policy classes $\Pi,\Gamma$ are $\Pi = \Pi_{\mrm{U}}(\cQ)$ and $\Gamma = \Gamma_{\mrm{U}}(\cP), \overline\Gamma_{\mrm{U}}(\cP)$ where $\mrm{U} = \mrm{H},\mrm{M},\mrm{S}$. This is the rigorous version of \eqref{eqn:simp:DRSC_value_func}. 
\par For simplicity, we write $v^*(x,\Pi,\Gamma):= v^*(\delta_{\set{x}},\Pi,\Gamma)$, and $v^*(\Pi,\Gamma)$ can be seen as a function $x\ra v^*(x,\Pi,\Gamma)$.

\subsection{Dynamic Programming}\label{a_sec:DPP}
\par In this section, we show that the solutions to the distributional robust Bellman equations \eqref{eqn:def_action_aware_DR_bellman_eqn} and \eqref{eqn:def_action_unaware_DR_bellman_eqn} exists under Assumption \ref{assump:cond_for_DPP}, and these solutions correspond to the DRSC value \eqref{eqn:DRSC_value_func}. 

Given the extended formulation in Appendix \ref{a_sec:DRSC_formulation}, we decompose and expand the statement of Theorem \ref{thm:main_body_dpp} into the following results:

\begin{proposition}\label{prop:solution_to_Bellman_eqn} Suppose Assumption \ref{assump:cond_for_DPP} is in force. 
The Bellman equations \eqref{eqn:def_action_aware_DR_bellman_eqn} and \eqref{eqn:def_action_unaware_DR_bellman_eqn} have unique points $u^*$ and $\bar u^*$, respectively within the class of uniformly continuous and non-negative functions $\X\ra\R$. Moreover, $\norm{u^*},\norm{\bar u ^*}\leq \beta r_\vee. $
\end{proposition}

\begin{theorem}[Dynamic Programming for CAA Adversaries]\label{thm:DPP_current_action_aware}
Suppose Assumption \ref{assump:cond_for_DPP} is in force, then $u^* = v^*(\Pi(\cQ),\Gamma(\cP))$ for each and every one of the 9 pairings $\Pi(\cQ) = \PiH(\cQ),\PiM(\cQ),\PiS(\cQ)$ and $\Gamma(\cP) =\GammaH(\cP),\GammaM(\cP),\GammaS(\cP)$. 
\end{theorem}

\begin{theorem}[Dynamic Programming for CAU Adversaries]\label{thm:DPP_current_action_unaware}
Suppose Assumption \ref{assump:cond_for_DPP} is in force, then there is a unique bounded continuous solution $\bar u^*$ to \eqref{eqn:def_action_unaware_DR_bellman_eqn}. Moreover, $\bar u^*$ is the optimal DRSC values
    \begin{align*}
    \bar u^* &= v^*(\PiH(\cQ),\barGammaH(\cP))\\
    &=  v^*(\PiM(\cQ),\barGammaH(\cP)) &&=  v^*(\PiM(\cQ),\barGammaM(\cP))\\
    &=  v^*(\PiS(\cQ),\barGammaH(\cP)) &&=  v^*(\PiS(\cQ),\barGammaM(\cP)) &&=  v^*(\PiS(\cQ),\barGammaS(\cP)).\\
    \end{align*}
\end{theorem}
\begin{remark} 
The equality $v^*(\PiH(\cQ),\barGammaH(\cP)) = v^*(\PiS(\cQ),\barGammaH(\cP))$ and $v^*(\PiM(\cQ),\barGammaM(\cP)) = v^*(\PiS(\cQ),\barGammaM(\cP))$ implies time-homogeneous (or stationary Markov) policies are optimal for history-dependent and Markov adversary.
\end{remark}

Clearly, Propositon \ref{prop:solution_to_Bellman_eqn}, Theorem \ref{thm:DPP_current_action_aware}, and \ref{thm:DPP_current_action_unaware} implies Theorem \ref{thm:main_body_dpp}. A history-dependent version of Theorem \ref{thm:DPP_current_action_aware} is established in \citet{gonzalez2002minimax}. In this paper, we will prove the more technically interesting Theorem \ref{thm:DPP_current_action_unaware}. The proof of the rest of the Theorem \ref{thm:DPP_current_action_aware} can be easily achieved by adapting the same proof techniques to deal with continuous state spaces in this paper to that of Theorem 2 from \citet{wang2023foundation}. 
\section{Proofs for Section \ref{sec:DPP} and \ref{a_sec:DPP}}

Let $U_b(\X)$ denote the space of uniformly bounded continuous functions on $\X$, which is a Banach space under the supremum norm $\norm{\cd}$.

\subsection{Proof of Proposition \ref{prop:solution_to_Bellman_eqn}}
We prove Proposition \ref{prop:solution_to_Bellman_eqn} by applying the Banach fixed-point to the mapping $\cT$ and $\overline\cT$. 
\begin{lemma}\label{lemma:contraction}
    Under Assumption \ref{assump:cond_for_DPP}, $\cT$ and $\overline\cT$ are $\alpha$-contractions on $(U_b(\X), \norm\cd)$; i.e. $\cT':U_b(\X)\ra U_b(\X)$ satisfies 
    $$\norm{\cT'(u_1) - \cT'(u_2)}\leq \alpha\norm{u_1-u_2}$$
    for all $u_1,u_2\in U_b(\X)$, where $\cT' =\cT,\overline\cT$. 
\end{lemma}
Therefore, there exists unique fixed-points $u^*$ for \eqref{eqn:def_action_aware_DR_bellman_eqn} and $\bar u^*$ \eqref{eqn:def_action_unaware_DR_bellman_eqn}. 
\par Moreover, for $\cT' =\cT,\overline\cT$ and $u' = u^*,\bar u^*$, we have that $$\norm{u'} =  \norm{\cT'(u') }\leq \norm r + \alpha \norm{u'} = r_\vee + \alpha \norm{u'}. $$ Hence, $\norm{u'}\leq \beta r_\vee$. 

\subsubsection{Proof of Lemma \ref{lemma:contraction}}
We will establish the result for $\overline\cT$, the statement for $\cT$ follows from the same arguments. First, we check that for $u\in U_b(\X)$, $\overline\cT(u)\in U_b(\X)$. Observe that by uniform continuity, for $x,z\in \X$ s.t. $d(x,z)\leq \epsilon$ there are $\delta,\delta',\delta'' > 0$ s.t.
\begin{align*}
&\abs{\overline\cT(u)(z) - \overline\cT(u)(x)}\\
&\leq\sup_{\phi\in\cQ}\abs{\inf_{\psi\in\cP}\int_{\A\times\W}r(x,a,w)  + \alpha u(f(x,a,w))\phi\times\psi(da,dw) + \sup_{\psi\in\cP}\int_{\A\times\W} - r(z,a,w)  - \alpha u(f(z ,a,w))\phi\times\psi(da,dw)} \\
&\stackrel{(i)}{\leq}\sup_{\phi\in\cQ}\sup_{\psi\in\cP}\int_{\A\times\W}|r(x,a,w)-r(z,a,w)|  + \alpha |u(f(x,a,w))- \alpha u(f(z,a,w))|\phi\times\psi (da,dw)\\
&\leq \sup_{a\in\A,w\in\W}|r(x,a,w)-r(z,a,w)| +\sup_{a\in\A,w\in\W}\sup_{y\in \bar B_{f(x,a,w)}(\delta')}\alpha |u(f(x,a,w))- u(y)|\\
&\leq \delta +\delta''
\end{align*}
uniformly in $x$, where $(i)$ follows from $|\inf f_1 + \sup f_2|\leq \max\set{|\sup (f_1+f_2)|,|\inf (f_1+f_2)|}\leq \sup |f_1+f_2|$ and $\bar B_{f(x,a,w)}(\delta'):= \set{y\in\X: d(f(x,a,w),y)\leq \delta'}$. Hence, $\overline\cT(u)\in U_b(\X)$. 
\par Next, we show that it is indeed a $\alpha$-contraction. Consider for $u_1,u_2\in U_b(\X)$, by the same argument, one has
\begin{align*}
\norm{\overline\cT(u_1) - \overline\cT(u_2)} 
 &\leq\sup_{x\in\X, \phi\in\cQ,\psi\in\cP}\int_{\A\times\W} \alpha |u_1(f(x,a,w))- u_2(f(x,a,w))|\phi\times\psi(da,dw)\\
  &\leq\sup_{x\in\X, a\in\A,w\in\W} \alpha |u_1(f(x,a,w))- u_2(f(x,a,w))|\\
  &\leq \alpha\norm{u_1-u_2}. 
\end{align*}
This completes the proof. 

\subsection{Proof of Theorem \ref{thm:DPP_current_action_unaware}}
We decompose our proof of Theorem \ref{thm:DPP_current_action_unaware} to two main Propositions as follows. 
\begin{proposition}\label{prop:general_unaware_adv_symmetric}
Under the assumptions of Theorem \ref{thm:DPP_current_action_unaware}, for any $\pi\in\Pi_{\mrm{U}}(\cQ)$, $$\inf_{\bar\gamma\in\overline{\Gamma}_{\mrm{U}}(\cP)}v(x,\pi,\bar\gamma)\leq \bar u^*(x),$$where $\mrm{U} = \mrm{H},\mrm{M},\mrm{S}$. 

\end{proposition}
In particular, Proposition \ref{prop:general_unaware_adv_symmetric} implies that $\bar u^*\geq  v^*(\PiH(\cQ),\barGammaH(\cP))$, $\bar u^*\geq  v^*(\PiM(\cQ),\barGammaM(\cP))$, and $\bar u^*\geq  v^*(\PiS(\cQ),\barGammaS(\cP))$.  

\begin{proposition}\label{prop:general_unaware_lb}
Under the assumptions of Theorem \ref{thm:DPP_current_action_unaware}, $$v^*(\PiS(\cQ),\barGammaH(\cP))\geq \bar u^*(x)$$
\end{proposition}
Therefore, we have that by the inclusion relationship $\PiH(\cQ)\supset \PiM(\cQ)\supset \PiS(\cQ)$, we have
\begin{align*}
    \bar u^*\geq  v^*(\PiH(\cQ),\barGammaH(\cP))\geq  v^*(\PiM(\cQ),\barGammaH(\cP)) \geq  v^*(\PiS(\cQ),\barGammaH(\cP)) \geq \bar u^*. 
\end{align*}
So all the quantities above are equal. Similarly, 
\begin{align*}
    u^*&=  v^*(\PiM(\cQ),\barGammaH(\cP)) &&\leq  v^*(\PiM(\cQ),\barGammaM(\cP))\leq u^*\\
    u^*&=  v^*(\PiS(\cQ),\barGammaH(\cP)) &&=  v^*(\PiS(\cQ),\barGammaM(\cP)) &&=  v^*(\PiS(\cQ),\barGammaS(\cP))\leq u^*.\\
    \end{align*}
This proves Theorem \ref{thm:DPP_current_action_unaware}. 
\subsubsection{Proof of Auxiliary Results for Theorem \ref{thm:DPP_current_action_unaware}}

\subsubsection{Proof of Proposition \ref{prop:general_unaware_adv_symmetric}}

Fix an arbitrary $\pi = (\pi_0,\pi_1,\ds)\in\Pi_{\mrm{U}}(\cQ)$. It suffice to show that for any $\epsilon > 0$ there exists $\bar\gamma\in\overline{\Gamma}_{\mrm{U}}(\cP)$ s.t. \begin{equation}\label{eqn:equiv_ineq_unaware_adv_symmetric}
v(x,\pi,\bar\gamma)\leq \bar u^*(x) +\epsilon.\end{equation}
\par Recall from \ref{prop:solution_to_Bellman_eqn} that $\norm{\bar u^*}\leq \beta r_\vee$. Define and denote the $T$-step truncated value with terminal reward $\bar u^*$ by 
\begin{equation}\label{eqn:truncated_val}
    v_T(x,\pi,\gamma):= E_x^{\pi,\gamma}\sqbk{\sum_{t=0}^{T_\epsilon-1}\alpha^t r(X_t,A_t,W_t) + \alpha^{T_\epsilon} \bar u^*(X_{T_\epsilon})}.
\end{equation}
Also, define $T_\eta = \ceil{\beta\log(2r_\vee\beta/\eta)}$ where $\beta = \frac{1}{1-\alpha}$. Then, $$\alpha^{T_\eta} \leq \crbk{1-\frac{1}{\beta}}^{\beta\log(2r_\vee\beta/\eta)}\leq \exp(-\log(2r_\vee\beta/\eta)) = \frac{\eta}{2r_\vee\beta}. $$
Thus, consider, for any $\pi\in \Pi_{\mrm{U}}(\cQ)$, $\gamma\in \Gamma_{\mrm{U}}(\cP)$, and $x\in\X$, we have \begin{equation}\label{eqn:truncated_val_bound}
 \begin{aligned}
\abs{v(x,\pi,\gamma ) - v_{T_\eta}(x,\pi,\gamma) } &=\abs{ E_x^{\pi,\gamma}\sqbk{\sum_{t=0}^\infty \alpha^t r(X_t,A_t,W_t)} - E_x^{\pi,\gamma}\sqbk{\sum_{t=0}^{T_\eta-1}\alpha^t r(X_t,A_t,W_t) + \alpha^{T_\eta} \bar u^*(X_{T_\eta})}}\\
&=\abs{ E_x^{\pi,\gamma}\sqbk{\sum_{t=T_\eta}^\infty \alpha^t r(X_t,A_t,W_t) - \alpha^{T_\eta}\bar u^*(X_{T_\eta})} }\\
&\leq\alpha^{T_\eta}\crbk{\abs{ E_x^{\pi,\gamma}\sum_{t=T_\eta}^\infty \alpha^{t-T_\eta} r(X_t,A_t,W_t)} + \abs{E_x^{\pi,\gamma} \bar u^*(X_{T_\eta})} }\\
&\leq \alpha^{T_\eta}\frac{2r_\vee}{1-\alpha}\\
&\leq \eta. 
\end{aligned}   
\end{equation}

Next, we focus on the history-dependent case $\mrm{U} = \mrm{H}$. Since $\bar u^*$ is the solution, we must have that for any $h_t\in\bd{H}_t$, 
$$\inf_{\psi\in\cP}\int_\W\int_\A r(x_t,a,w) + \alpha \bar u^*(f(x_t,a,w))\pi_t(da|h_t)\psi(dw) \leq \bar u^*(x_t)$$
Fix any $\delta > 0$. Let $\Psi^\delta_t(h_t)$ be a set of $\psi\in\cP$ s.t. 
\begin{equation}\label{eqn:pf_thm1_select_hd_adv_delta}
 \int_\W\int_\A r(x_t,a,w) + \alpha \bar u^*(f(x_t,a,w))\pi_t(da|h_t)\psi(dw) \leq \bar u^*(x_t) + \delta.   
\end{equation}
We want to show that there is a $\bd H_t\ra \cP(\cW)$ measurable selection $\bar\gamma_t^\delta(dw|h_t)\in\Psi_t^\delta(h_t)$. 

We note that  $$\set{\psi\in\cP:\int_\W\int_\A r(x_t,a,w) + \alpha \bar u^*(f(x_t,a,w))\pi_t(da|h_t)\psi(dw) \leq \bar u^*(x_t) + \delta}$$ is a closed set. It is clearly nonempty, as $\bar u^*$ satisfies \eqref{eqn:def_action_unaware_DR_bellman_eqn}. 
Now, by the Kuratowski-Ryll-Nardzewski measurable selection theorem, we show that for any open set $\cD$ of $\cP$, we have that $$\set{h_t\in\bd H_t:\varnothing\neq\set{\psi\in\cP:\int_\W\int_\A r(x_t,a,w) + \alpha \bar u^*(f(x_t,a,w))\pi_t(da|h_t)\psi(dw) \leq \bar u^*(x_t) + \delta}\cap\cD}\in\cB(\bd H_t). $$
Note that, 
$$\begin{aligned}
&\varnothing\neq\set{\psi\in\cP:\int_\W\int_\A r(x_t,a) + \alpha \bar u^*(f(x_t,a,w))\pi_t(da|h_t)\psi(dw) \leq \bar u^*(x_t) + \delta}\cap\cD\\
\iff&\exists \psi\in\cD: \int_\W\int_\A r(x_t,a,w) + \alpha \bar u^*(f(x_t,a,w))\pi_t(da|h_t)\psi(dw) \leq \bar u^*(x_t) + \delta\\
\iff&\inf_{\psi\in\cD}\int_\W\int_\A r(x_t,a,w) + \alpha \bar u^*(f(x_t,a,w))\pi_t(da|h_t)\psi(dw) - \bar u^*(x_t) - \delta\leq 0
\end{aligned}$$
Recall that $\cP(\cW)$ is endowed with the Lévy–Prokhorov metric. Since $\W$ is separable, so is $\cP(\cW)$. Observe that by bounded convergence and the continuity of $f$ and $\bar u^*$, for $w_k\ra w$ on $\W$, we have that $$\lim_{k\ra\infty}\int_\A r(x_t,a,w) + \alpha \bar u^*(f(x_t,a,w_k))\pi_t(da|h_t) = \int_\A r(x_t,a,w) + \alpha \bar u^*(f(x_t,a,w))\pi_t(da|h_t)$$ i.e. $w\ra \int_\A r(x_t,a,w) + \alpha \bar u^*(f(x_t,a,w_k))\pi_t(da|h_t)$ is bounded continuous. Hence,  
$$\psi\ra\int_\W\int_\A r(x_t,a,w) + \alpha \bar u^*(f(x_t,a,w))\pi_t(da|h_t)\psi(dw)$$is continuous. Therefore, we have that $$c(h_t):= \inf_{\psi\in\cD}\int_\W\int_\A r(x_t,a,w) + \alpha \bar u^*(f(x_t,a,w))\pi_t(da|h_t)\psi(dw) - \bar u^*(x_t) - \delta$$is $\cB(\bd H_t)\ra\R$ measurable, by the measurability of $\pi_t$ and that the infimum can be taken over a dense subset. Hence, the sub-level set $c(h_t)\leq 0$ is $\cB(\bd H_t)$ measurable. 

Therefore, the measurable selection theorem applies and we conclude that there exists $\bar\gamma_t^\delta:\bd{H}_t\ra \cP$ measurable s.t. 
$$\int_\W\int_\A r(x_t,a,w) + \alpha \bar u^*(f(x_t,a,w))\pi_t(da|h_t)\bar\gamma^\delta_t(dw|h_t) \leq \bar u^*(x_t) + \delta.$$
Since this can be done for each $t$, we can construct $\bar\gamma^\delta\in\barGammaH(\cP)$ s.t. the above inequality holds for each and every $t$. 
\par Now, we first consider $$
\begin{aligned}
E_x^{\pi,\bar\gamma^\delta} [r(X_t,A_t,W_t) + \alpha \bar u^*(X_{t+1})] &= E_x^{\pi,\bar\gamma^\delta} [r(X_t,A_t,W_t) + \alpha \bar u^*(f(X_{t},A_t,W_t))]\\
&= E_x^{\pi,\bar\gamma^\delta}  E_x^{\pi,\bar\gamma^\delta}[r(X_t,A_t.W_t) + \alpha \bar u^*(f(X_{t},A_t,W_t))|H_t]\\
&= E_x^{\pi,\bar\gamma^\delta}  \int_\W\int_\A r(X_t,a,w) + \alpha \bar u^*(f(X_{t},a,w))\pi_t(da|H_t)\bar\gamma^\delta_t(dw|H_t)\\
&\leq E_x^{\pi,\bar\gamma^\delta}  \bar u^*(X_t) + \delta. 
\end{aligned}$$
Recall the definition \eqref{eqn:truncated_val}, we have that for any $T\geq1$, \begin{equation}\label{eqn:truncated_sum_bound}\begin{aligned}
v_{T}(x,\pi,\bar\gamma^\delta) &=  E_x^{\pi,\bar\gamma^\delta}\sqbk{\sum_{t=0}^{T-1}\alpha^t r(X_t,A_t.W_t) + \alpha^{T} \bar u^*(X_{T})}\\
&= E_x^{\pi,\bar\gamma^\delta}\sqbk{\sum_{t=0}^{T-2}\alpha^t r(X_t,A_t,W_t)}  + \alpha^{T-1}E_x^{\pi,\bar\gamma^\delta}\sqbk{ r(X_{T-1},A_{T-1},W_{T-1}) +\alpha \bar u^*(X_{T})}\\
&\leq E_x^{\pi,\bar\gamma^\delta}\sqbk{\sum_{t=0}^{T-2}\alpha^t r(X_t,A_t,W_t)}  + \alpha^{T-1}E_x^{\pi,\bar\gamma^\delta}\sqbk{ \bar u^*(X_{T-1})} + \alpha^{T-1}\delta\\
&\leq v_{T-1}(x,\pi,\bar\gamma^\delta)+ \alpha^{T-1}\delta. 
\end{aligned}\end{equation}
Therefore, by induction on $T$, we conclude that 
$$\begin{aligned}
v_{T}(x,\pi,\bar\gamma^\delta) &\leq v_{1}(x,\pi,\bar\gamma^\delta) +\delta\sum_{t=1}^{T-1}\alpha^{t}\\
&\leq E_x^{\pi,\bar\gamma^\delta} [r(X_0,A_0,W_0) + \alpha \bar u^*(X_{1})] -\delta +\beta\delta\\
&\leq \bar u^*(x) + \beta \delta.
\end{aligned}$$
\par Since $\delta$ is arbitrary, choosing $\delta = \epsilon/(2\beta)$ and $T = T_{\epsilon/2}$, we conclude that by \eqref{eqn:truncated_val_bound}, $$\begin{aligned}
v(x,\pi,\bar\gamma^\delta ) &\leq v_{T_{\epsilon/2}}(x,\pi,\bar\gamma^\delta) + \frac{\epsilon}{2}\\
&\leq \bar u^*(x) + \beta\delta + \frac{\epsilon}{2}\\
&\leq \bar u^*(x) + \epsilon;
\end{aligned}$$
i.e. inequality \eqref{eqn:equiv_ineq_unaware_adv_symmetric} holds with adversarial policy $\bar\gamma^\delta$. This completes the proof for the case $\mrm{U} = \mrm{H}$. 
\par For cases $\mrm{U} = \mrm{M},\mrm{S}$, the proof remains the same except we choose the adversary to be Markov or time-homogeneous, in the presence of a Markov or time-homogeneous controller, respectively. For instance, in the time-homogeneous case, given any $\pi(da|x)$ we choose a policy $\bar\gamma^{\delta}\in \barGammaS(\cP)$ s.t.
$$ \int_\W\int_\A r(x_t,a,w) + \alpha \bar u^*(f(x_t,a,w))\pi(da|x_t)\bar\gamma^{\delta}(dw|x_t) \leq \bar u^*(x_t) + \delta.$$
for every $x_t$. A measurable choice is always possible because $\bar\gamma^\delta(da|x_t)$ has the same information dependence on $x_t$ as $\pi(da|x_t)$.

\subsubsection{Proof of Proposition \ref{prop:general_unaware_lb}}

\par Since $\bar u^*$ is the solution to \eqref{eqn:def_action_unaware_DR_bellman_eqn}, by the same measurable selection argument and separability, for any fixed $\delta > 0$, there exists measurable $\pi^\delta(da|x)$ s.t. $\pi^\delta(da|x)\in\cQ$ for all $x\in\X$ and
\begin{equation}\label{eqn:bellman_sup_pi_delta}
\bar u^*(s)\leq \inf_{\psi\in\cP}\int_{\A}\int_{\W}r(x,a,w) + \alpha  \bar u^*(f(x,a,w))\pi^\delta(da|x)\psi(dw) + \delta.    
\end{equation}  
Let $\pi^\delta = (\pi^\delta,\pi^\delta,\ds)\in\PiS(\cQ)$. We consider for any $\bar\gamma = (\bar\gamma_0,\bar\gamma_1,\ds)\in \barGammaH(\cP)$, 
\begin{align*}
 E_\mu^{\pi^\delta,\bar\gamma}\sqbk{ u^*(X_{t+1})} &=  E_\mu^{\pi^\delta,\bar\gamma}E_\mu^{\pi^\delta,\bar\gamma}\sqbkcond{ \bar u^*(f(X_t,A_t,W_t))}{G_{t}} \\
&= E_\mu^{\pi^\delta,\bar\gamma}  \int_{\W} \bar u^*(f(X_t,A_t,w))\bar \gamma_t(dw|G_{t})\\
&= E_\mu^{\pi^\delta,\bar\gamma}  \int_{\A}\int_{\W} \bar u^*(f(X_t,a,w))\bar \gamma_t(dw|H_{t},a)\pi^\delta(da|X_t)\\
&= E_\mu^{\pi^\delta,\bar\gamma}  \int_{\A}\int_{\W} \bar u^*(f(X_t,a,w))\bar \gamma_t(dw|G_{t-1},X_t,a)\pi^\delta(da|X_t)
\end{align*}
Given $G_{t-1}$ and $X_t$, 
\begin{align*}\int_{\A}\int_{\W} \bar u^*(f(X_t,a,w))\bar \gamma_t(dw|G_{t-1},X_t,a)\pi^\delta(da|X_t)&\geq \inf_{\psi\in\cP}\int_{\A}\int_{\W} \bar u^*(f(X_t,a,w))\psi(dw)\pi^\delta(da|X_t)\\
\end{align*}
Therefore, by \eqref{eqn:bellman_sup_pi_delta}
\begin{align*}
 &E_\mu^{\pi^\delta,\bar\gamma}\sqbk{r(X_{t}, A_t,W_t) + \alpha u^*(X_{t+1})} \\
 &\geq E_\mu^{\pi^\delta,\bar\gamma} \sqbk{\int_{\A}r(X_t,a,w)\pi^{\delta}(da|X_t) +  \alpha\inf_{\psi\in\cP}\int_{\A}\int_{\W} \bar u^*(f(X_t,a,w))\psi(dw)\pi^\delta(da|X_t)}\\
 &= E_\mu^{\pi^\delta,\bar\gamma} \sqbk{  \inf_{\psi\in\cP}\int_{\A}\int_{\W} r(X_t,a,w) + \alpha\bar u^*(f(X_t,a,w))\psi(dw)\pi^\delta(da|X_t)}\\
 &\geq E_\mu^{\pi^\delta,\bar\gamma}\bar u^*(X_t)-\delta.
 \end{align*}
By the same technique as in \eqref{eqn:truncated_sum_bound}, we see that for all $T$, $$v_{T}(x,\pi^\delta,\bar\gamma)\geq \bar u^*(x) - \beta\delta$$uniformly in $x$. 
\par Since $\pi\in\PiS(\cQ)$,
\begin{align*}
v^*(\PiS(\cQ),\barGammaH(\cP))&\geq \inf_{\bar\gamma\in\barGammaH(\cP)} v(\pi^\delta,\bar\gamma)\\
&\geq \inf_{\bar\gamma\in\barGammaH(\cP)} v_{T_\epsilon}(\pi^\delta,\bar\gamma)-\epsilon\\
&\geq\bar u^* - \beta\delta-\epsilon\\
\end{align*}
by \eqref{eqn:truncated_val_bound}. Since $\epsilon$ and $\delta$ are arbitrary, we complete the proof.
\section{Proofs for Section \ref{sec:UB}}
\subsection{Proof of Proposition \ref{prop:bound_est_err_by_oprtr_gap}}
Define a sequence of functions $u_0 \equiv 0$ and $u_{k+1} = \bd{T}'(u_{k})$. By the Banach fixed point theorem, $u_k\ra \hat u$ in uniform norm. Since $u' = \cT'(u')$, the error \begin{align*}\Delta_{k+1}&:=u_{k+1}-u'\\
&= \bd{T}'(u_{k}) - \bd{T}'(u') +  \bd{T}' (u') - \cT' (u')\\
&= [\bd{T}'(u'+\Delta_{k})  - \bd{T}'(u')] + [  \bd{T}' (u') - \cT' (u')]\\
&=:\bd{H}(\Delta_{k})+U.
\end{align*}
By Proposition \ref{prop:solution_to_Bellman_eqn}, it is easy to see that $\bd{H}$ is also a $\alpha$-contraction on $U_b(\X)$, with $0$ as its unique fixed point. 
\par We claim that for $k\geq 1$, 
$$\norm{\Delta_k}\leq \beta \alpha^{k-1} + \sum_{j=0}^{k-1}\alpha^j\norm U. $$
We check this by induction: for $k = 1$, 
$$\begin{aligned}
    \norm{\Delta_1} &\leq \norm{\bd{H}(\Delta_{0})}+\norm{U}\\
    &= \norm{\bd{H}(\Delta_{0}) - \bd{H}(0)}+\norm{U}\\
    &\leq \alpha \norm{u'}+\norm{U}\\
    &\leq \alpha\beta + \norm{U}. 
\end{aligned}$$
For the induction step, we have that 
$$\begin{aligned}
    \norm{\Delta_{k+1}} &\leq \norm{\bd{H}(\Delta_{k})- \bd{H}(0)}+\norm{U}\\
    &\leq \alpha \norm{\Delta_k}+\norm{U}\\
    &\leq\beta \alpha^k + \sum_{j=0}^{k}\alpha^j\norm{U}. 
\end{aligned}$$
where the last inequality follows from the induction assumption. 
\par Therefore, \begin{align*}\norm {\hat u - u'} = \lim_{k\ra\infty}\norm{\Delta_k} \leq  \sum_{j=0}^\infty \alpha^j\norm{U} = \beta\norm{\bd{T}' (u') - \cT' (u')}. \end{align*}
\subsection{Proof of Theorem \ref{thm:aware_Wd_sample_complexity}}\label{a_sec:proof:thm:aware_Wd_sample_complexity}
\par We remark that our proof techniques have similarities with that in \citet{lee2018minimax}. By Proposition \ref{prop:bound_est_err_by_oprtr_gap}, to achieve an upper bound on the uniform learning error, it suffices to prove an upper bound for $\norm{\bd{T} (u^*) - \cT (u^*)}$.
\par To do this, we first rewrite the Bellman operator using its dual form. By strong duality \citep{blanchet2019quantifying},  for $\cP = \set{\mu\in\cP(\cX):W(\mu,\mu_0)\leq \delta}$
$$\inf_{\psi\in \cP}\int_{\W} [r(z,w)+\alpha u^*(f(z,w))]\psi(dw) = \sup_{\lambda\geq 0}-\lambda\delta + \int_{\W}\inf_{y\in \W} \sqbk{ r(z,y)+\alpha u^*( f(z,y))  + \lambda c(w,y)}\mu_0(dw).$$
\par Notice that since $$\int_{\W}\inf_{y\in \W} \sqbk{r(z,y)+\alpha u^*( f(z,y))  + \lambda c(w,y)}\mu_0(dw)\leq \int_{\W} [r(z,w)+\alpha u^*( f(z,w))]\mu_0(dw) \leq 1+\alpha \norm{u^*}$$
and $$\inf_{\psi\in \cP}\int_{\W} r(z,w) + \alpha u^*(f(z,w))\psi(dw) \geq 0,$$
it suffices to maximize $\lambda$ over $\Lambda := [0,\delta\inv (1+\alpha \norm{u^*})]$. 
\par Therefore, we have that $$\begin{aligned}
&\norm{\bd{T} (u^*) - \cT (u^*)}\\
&\leq \sup_{z\in\X\times \A} \abs{ \sup_{\lambda\in\Lambda}\sqbk{ \int_{\W}\inf_{y\in \W} \sqbk{ r(z,y) + \alpha u^*(f(z,y))  + \lambda c(w,y)}\mu_0 (dw)-\lambda\delta } -\sup_{\lambda\in\Lambda}\sqbk{\int_{\W}\inf_{y\in \W} \sqbk{ r(z,y) + \alpha u^*(f(z,y))  + \lambda c(w,y)}\hat\mu(dw)-\lambda\delta }} \\
&\leq \sup_{z\in\X\times \A}\sup_{\lambda\in\Lambda} \abs{ \int_{\X}\inf_{y\in \W} \sqbk{ r(z,y) + \alpha u^*(f(z,y))  + \lambda c(w,y)}(\mu_0 - \hat\mu)(dw)}\\
& = \sup_{\theta\in\Theta}\abs{(\mu_0-\hat \mu)[g_\theta]}
\end{aligned}$$
where $\Theta = \set{\theta = (z,\lambda): z\in\X\times \A,  \lambda\in\Lambda}$ and $$g_\theta = \inf_{y\in \W} \sqbk{ r(z,y) + \alpha u^*(f(z,y)) + \lambda c(\cd,y)}. $$ 
Therefore, the estimation error is bounded by a supremum of empirical process.
\par To bound this, we use the Rademacher process and a chaining argument. Specifically, for fixed sequence $\bd{w}:=\set{w_i\in\W:i=1,2,\ds, n}$, we define the Rademacher process indexed by $g_\theta\in\cG$ as
\begin{equation}\label{eqn:Wd_aware_Rad_process}R_n(\bd{w},g_\theta):= \frac{1}{\sqrt{n}}\sum_{i=1}^n \epsilon_i g_\theta(w_i)= \frac{1}{\sqrt{n}}\sum_{i=1}^n \epsilon_i \inf_{y\in \W}\sqbk{r(z,y) + \alpha u^*(f(z,y)) + \lambda c(w_i,y)}.
\end{equation} The empirical and population Rademacher complexities of the function class $\cG:= \set{g_\theta:\theta\in\Theta}$
\begin{equation}\label{eqn:Wd_aware_Rad_cplx}
 \cR_n(\bd{w},\cG) := E_\epsilon\sup_{g\in\cG} \frac{1}{\sqrt{n}}R_n(\bd{w},g), \quad\text{and}\quad 
 \cR_n(\cG) := E_{\mu_0^n}\cR_n(\bd{W},\cG)
\end{equation}
where $\mu_0^n = \mu_0\times\ds\times\mu_0$ the $n$-fold product measure, and $W = (W_1,\ds,W_n)$. 
\par From empirical process theory, see e.g. \citet[Theorem 4.10]{wainwright2019high_stats}, w.p. at least $1-\eta$, 
\begin{equation}\label{eqn:bound_emp_process_by_Rad_cplx}
    \sup_{g\in\cG} \abs{(\mu_n-\mu_0)[g]}\leq  2 \cR_n(\cG)+ \sqrt{\frac{2g_\vee}{n}\log\crbk{\frac{1}{\eta}}}.
\end{equation}
where $g_\vee:=\sup_{g\in\cG}\|g\| \leq 1+\alpha \norm{u^*}\leq 1+\alpha\beta = \beta$. Thus, we proceed to bound $\cR_n(\bd{w},\cG)$ and hence $\cR_n(\cG)$. We achieve this by using subgaussian processes and entropy integrals. 
\par Specifically, we consider the moment generating function of the Rademacher process \eqref{eqn:Wd_aware_Rad_process}. For $\xi$ in some neighborhood of the origin $$\begin{aligned}
&E_\epsilon \exp[\xi(R_n(\bd{w},g_\theta) - R_n(\bd{w},g_{\theta'}))] \\
&= E_{\epsilon}\exp\crbk{\frac{\xi}{\sqrt{n}}\sum_{i=1}^n \epsilon_i \sqbk{\inf_{y\in \W}\sqbk{r(z,y)+\alpha u^*(f(z,y)) + \lambda c(w_i,y)}  -  \inf_{y\in \W}\sqbk{r(z',y)+\alpha u^*(f(z',y)) + \lambda' c(w_i,y)} } }\\
&\leq \exp\crbk{\frac{\xi^2}{2n}\sum_{i=1}^n \sqbk{ \inf_{y\in \W}\sqbk{r(z,y)+\alpha u^*(f(z,y)) + \lambda c(w_i,y)}  -  \inf_{y\in \W}\sqbk{r(z',y)+\alpha u^*(f(z',y)) + \lambda' c(w_i,y)}}^2}\\
&\leq\exp\crbk{\frac{\xi^2}{2n}\sum_{i=1}^n  \sup_{y\in \W}\abs{r(z,y)+\alpha u^*(f(z,y)) - r(z',y) -\alpha u^*(f(z',y))  + (\lambda'-\lambda)c(w_i,y)}^2 }\\
&\stackrel{(i)}{\leq}\exp\crbk{\frac{\xi^2}{2} \crbk{\sup_{y\in \W}\abs{ r(z,y)+\alpha u^*(f(z,y)) - r(z',y)-\alpha u^*(f(z',y))}+ \abs{\lambda- \lambda'} c_\vee}^2}\\
&\stackrel{(ii)}{\leq}\exp\crbk{\frac{\xi^2}{2} \crbk{L(\abs{ x-x'}+|a-a'|)+ \abs{\lambda- \lambda'} c_\vee}^2}\\
\end{aligned}$$
where $(i)$ uses the transport cost being bounded by $c_\vee$ and $(ii)$ follows from the uniform Lipschitzness in Assumption \ref{assump:aware_lip_val_bounded_sps}. Therefore, defining $$\rho(\theta,\theta') := L(|x-x'| +|a-a'|) +  c_\vee|\lambda-\lambda'|,$$ which is a distance on $\Theta$, we obtain that
$$E_\epsilon \exp[\xi(R_n(\bd{w},g_{\theta}) - R_n(\bd{w},g_{\theta'}))]\leq \exp\crbk{\frac{\xi^2}{2}\rho^2(g_{\theta},g_{\theta'})}. $$
This shows that the stochastic process $\set{R_n(\bd{w},g_\theta):\theta\in\Theta}$ is subgaussian w.r.t. $\rho$. 
\par Therefore, using Dudley’s entropy integral for subgaussian processes \citep[Chapter 5]{wainwright2019high_stats}, the empirical Rademacher complexity in \eqref{eqn:Wd_aware_Rad_cplx} can be bounded by \begin{equation}\label{eqn:Wd_aware_entropy_int}
\begin{aligned}
\cR_n(\bd{w},\cG)
&= E_\epsilon\sup_{\theta\in\Theta} \frac{1}{\sqrt{n}}R_n(\bd{w},g_\theta)\\
&\leq \frac{32}{\sqrt{n}}\int_0^{D_\vee}\sqrt{\log\cN(\epsilon;\Theta,\rho)} d\epsilon
\end{aligned}    
\end{equation}
w.p.1., where $\cN(\epsilon;\Theta,\rho)$ is the $\epsilon$ covering number of $\Theta$ in distance $\rho$ and $$\begin{aligned}
D_\vee &:=   L(\diam(\X) + \diam( \A)) + c_\vee \delta\inv\beta +1\\
&\geq L(\diam(\X) + \diam( \A)) + c_\vee\diam(\Lambda) +1\\
&\geq \sup_{\theta,\theta'\in\Theta}\rho(\theta,\theta')  
\end{aligned}$$ is an upper bound on the diameter of $\Theta$ in terms of $\rho$.
\par Note that as the r.h.s. of \eqref{eqn:Wd_aware_entropy_int} is deterministic, we take expectation over $\bd{W}\sim \mu^n_0$ to conclude that the population Rademacher complexity $$\cR_n(\cG)= E_{\mu^n_0}\cR_n(\bd{W},\cG)\leq \frac{32}{\sqrt{n}}\int_0^{D_\vee}\sqrt{\log\cN(\epsilon;\Theta,\rho)} d\epsilon$$satisfying the same bound. 
Moreover, the covering number $$\begin{aligned}
\cN(\epsilon;\Theta,\rho) &= \cN(\epsilon;\X\times\A,L|\cd|)\cd\cN(\epsilon;\Lambda,c_\vee|\cd|)\\
&= \cN(\epsilon;\X,L|\cd|)\cd \cN(\epsilon;\A,L|\cd|)\cd\cN(\epsilon;\Lambda,c_\vee|\cd|)\\
&\leq \crbk{1+\frac{L\diam(\X)}{\epsilon}}^{d_\X} \crbk{1+\frac{L\diam(\A)}{\epsilon}}^{d_\A}\crbk{1+ \frac{c_\vee\diam(\Lambda)}{\epsilon}}.
\end{aligned}$$
Therefore, the entropy integral $$\begin{aligned}
\int_0^{D_\vee}\sqrt{\log\cN(\epsilon;\Theta,\rho)} d\epsilon &\leq \int_0^{D_\vee}\sqrt{ (d_{\X}+d_{\A}+ 1)\log\crbk{1+\max\set{\frac{L)\diam(\A)}{\epsilon}, \frac{L\diam(\X)}{\epsilon}, \frac{c_\vee\diam(\Lambda)}{\epsilon}} }}d\epsilon\\
&\leq \int_0^{D_\vee}\sqrt{ (d_{\X}+d_{\A}+ 1)\log(D_\vee/\epsilon)} d\epsilon\\
& = \frac{\sqrt{\pi}}{2} D_\vee\sqrt{d_{\X}+d_{\A}+ 1}
\end{aligned}$$
\par We conclude that by Proposition \ref{prop:bound_est_err_by_oprtr_gap} and \eqref{eqn:bound_emp_process_by_Rad_cplx}, the estimation error $$\begin{aligned}
\norm{\hat u - u^*} &\leq \beta \norm{\bd{T} (u^*) (x)- \cT (u^*)(x) } \\
&\leq \beta\sup_{\theta\in\Theta} \abs{(\mu_n-\mu_0)[g_\theta]}\\
&\stackrel{(i)}{\leq}  2\beta  \cR_n(\cG)+ \beta\sqrt{\frac{2g_\vee}{n}\log\crbk{\frac{1}{\eta}}}\\
&\leq \crbk{32\sqrt{\pi} \beta D_\vee\sqrt{d_{\X}+d_{\A}+1}+ \sqrt{2}\beta^{3/2}\sqrt{\log\crbk{\frac{1}{\eta}}}}n^{-\frac{1}{2}}
\end{aligned}$$where $(i)$ holds w.p. at least $1-\eta$. This bound implies the theorem as we note that $d_\X\geq 1$.

\subsection{Proof of Theorem \ref{thm:aware_fk_sample_complexity}}
As in the previous proof, we bound 
$$\norm{\bd{T}(u^*) - \cT(u^*)} \leq \sup_{z\in \X\times\A}\abs{ \inf_{\psi\in \cP}\int_{\W}[r(z,w)+\alpha u^*(f(z,w))]\psi(dw)- \inf_{\psi\in\widehat \cP}\int_{\W}[r(z,w) + \alpha u^*(f(z,w))]\psi(dw) }$$
Define the function class $\cF := \set{r(z,\cd)+\alpha u^*(f(z,\cd)): z\in \X\times \A)}$. By \citet[Corollary 1]{duchi2021learning}, the r.h.s. satisfies w.p. at least $1-2\cN(\epsilon/3;\cF,\norm{\cd})e^{-t}$, 
$$ \sup_{z\in \X\times\A}\abs{ \inf_{\psi\in \cP}\int_{\W} [r(z,w) + \alpha u^*(f(z,w))]\psi(dw)- \inf_{\psi\in\widehat \cP}\int_{\W} [r(z,w) + \alpha u^*(f(z,w))]\psi(dw) }\leq 30 (1+\alpha\beta) \epsilon$$
where $$\epsilon = n^{-\frac{1}{k'\vee 2}}c_k(\delta)^2\crbk{\frac{c_k(\delta)}{c_k(\delta)-1}\vee 2}\crbk{\frac{1}{k}+\sqrt{t+2\log n}}.$$
Therefore, choosing $t = \log(2\cN(\epsilon/3;\cF,\norm{\cd})/\eta)$, we have that w.p. at least $1-\eta$
$$\norm{\bd{T}(u^*) - \cT(u^*)} \leq 30(1+\alpha\beta) n^{-\frac{1}{k'\vee 2}}c_k(\delta)^2\crbk{\frac{c_k(\delta)}{c_k(\delta)-1}\vee 2}\crbk{\frac{1}{k}+\sqrt{\log(2\cN(\epsilon/3;\cF,\norm{\cd})+\log\frac{1}{\eta}+2\log n}}.$$
By the uniform Lipschitz assumptions and that $\epsilon\geq n^{-1/2}$ for all $t\geq 1$, we have that by \citet[Chapter 2.7.4]{van_der_vaart1996week_conv_and_emprical_meas}
$$ \begin{aligned}
\log \cN(\epsilon/3;\cF,\norm{\cd}) &\leq \log \cN\crbk{\frac{\epsilon\wedge 1}{3L};\X\times\A,\abs{\cd}} \\ 
&\leq d_{\X}\log \crbk{1+\frac{ 3L\diam(\X)}{\epsilon\wedge 1}}+{d_{\A}}\log\crbk{1+\frac{3L\diam(\A)}{\epsilon\wedge 1}}\\
&\leq d_{\X}\log \crbk{1+3L\diam(\X)}+{d_{\A}}\log\crbk{1+3L\diam(\A)} + \frac{1}{2}(d_\X +d_{\A})\log n.
\end{aligned}$$ 
Therefore, defining $D = d_{\X}\log \crbk{1+3L\diam(\X)}+{d_{\A}}\log\crbk{1+3L\diam(\A)}$, we conclude that by Proposition \ref{prop:bound_est_err_by_oprtr_gap}, the estimation error $$\begin{aligned}
\norm{\hat u - u^*} &\leq \beta \norm{\bd{T} (u^*) (x)- \cT (u^*)(x) } \\
&\stackrel{(i)}{\leq} 30\beta^2 n^{-\frac{1}{k'\vee 2}}c_k(\delta)^2\crbk{\frac{c_k(\delta)}{c_k(\delta)-1}\vee 2}\crbk{\frac{1}{k} + \sqrt{D+\log\frac{1}{\eta} + 2(d_\X+ d_\A)\log n}} 
\end{aligned}$$
where $(i)$ holds w.p. at least $1-\eta$.

\subsection{Proof of Theorem \ref{thm:unaware_Wd_sample_complexity}}\label{a_sec:proof:thm:unaware_Wd_sample_complexity}
In this proof, we first consider general Polish action space and then specialize to finite action space to achieve $n^{-1/2}$ rate. Through the proof, we identify possible structures of the controller's decision space $\cQ$ so that 
\par We employ the same proof strategy as that of Theorem \ref{thm:aware_Wd_sample_complexity} in Appendix \ref{a_sec:proof:thm:aware_Wd_sample_complexity}. By the strong duality, positivity, and Bellman equation \eqref{eqn:def_action_unaware_DR_bellman_eqn}, we have that
$$\begin{aligned}\bar u^*(x)&= \sup_{\phi\in\cQ}\int_{\A}  \inf_{\psi\in \cP}\int_{\W} [r(x,a,w) +\alpha \bar u^* (f(x,a,w))]\psi(dw)\phi(da)\\ 
&= \sup_{\phi\in\cQ}\int_{\A} \sup_{\lambda\geq 0}-\lambda\delta + \int_{\W}\inf_{y\in \W} \sqbk{ [r(x,a,y) +\alpha \bar u^*( f(x,a,y))] +\lambda c(w,y)}\mu_0(dw)\phi(da)\end{aligned}$$
By the same argument, the supremum is achieved within $\Lambda:= [0,\delta\inv(1+\alpha \norm{\bar u^*})]$. So, we have that
$$ \begin{aligned}\norm{\overline{\bd{T}} (\bar u^*) (x)- \cT (\bar u^*)(x) } &\leq \sup_{x\in  \X,\phi\in\cQ}\abs{ \inf_{\psi\in \widehat \cP}\int_{\W} \int_{\A} [r(x,a,w) + \alpha \bar u^* (f(x,a,w))]\phi(da)\psi(dw)- \inf_{\psi\in \cP}\int_{\W} \int_{\A} [r(x,a,w) + \alpha \bar u^* (f(x,a,w))]\phi(da)\psi(dw)}\\
&\leq \sup_{x\in  \X,\phi\in\cQ,\lambda\in\Lambda}\abs{ \int_{\W} \inf_{y\in \W} \sqbk{ \int_{\A} [r(x,a,y) + \alpha \bar u^*( f(x,a,y))] \phi(da) + \lambda c(w,y)}(\mu_0-\hat \mu)(dw)}\\
&=:\sup_{g\in\cG}\abs{(\mu_0-\hat \mu)[g]}
\end{aligned}$$
Here, the parametric function class $\cG$ is characterized by $(x,\psi,\lambda)\in\Theta = \X\times\cQ\times\Lambda$ and $$\cG:= \set{w\ra \inf_{y\in \W} \sqbk{ \int_{\A} [r(x,a,y)+ \alpha \bar u^*( f(x,a,y))] \phi(da) + \lambda c(w,y)}
:(x,\psi,\lambda)\in\Theta }.$$
\par To bound the previous empirical process supremum, we still employ the Rademacher complexity bound as in \eqref{eqn:bound_emp_process_by_Rad_cplx}. In this case, for $g\in\cG$ and sequence $\bd{w}:=\set{w_i\in\W:i=1,2,\ds, n}$, the Rademacher process is $$R_n(\bd{w},g):= \frac{1}{\sqrt{n}}\sum_{i=1}^n \epsilon_i g(w_i)= \frac{1}{\sqrt{n}}\sum_{i=1}^n \epsilon_i \inf_{y\in \W} \sqbk{ \int_{\A} [r(x,a,y) + \alpha \bar u^*( f(x,a,y))] \phi(da) + \lambda c(w_i,y)},
$$
compare to \eqref{eqn:Wd_aware_Rad_process}, and the empirical and population complexities are defined as in \eqref{eqn:Wd_aware_Rad_cplx} accordingly. 
We then consider the moment generating function: for $\xi$ in some neighborhood of the origin $$\begin{aligned}
&E_\epsilon \exp[\xi(R_n(\bd{w},g_\theta) - R_n(\bd{w},g_{\theta'}))] \\
&= E_{\epsilon}\exp\crbk{\frac{\xi}{\sqrt{n}}\sum_{i=1}^n \epsilon_i [g_\theta(w_i)-g_{\theta'}(w_i)]}\\
&\leq \exp\crbk{\frac{\xi^2}{2n}\sum_{i=1}^n[g_\theta(w_i)-g_{\theta'}(w_i)]^2}\\
&\leq\exp\crbk{\frac{\xi^2}{2n}\sum_{i=1}^n  \sup_{y\in \W}\abs{\int_\A [r(x,a,y) + \alpha \bar u^*(f(x,a,y))]\phi(da)  - \int_\A [r(x',a,y) + \alpha \bar u^*( f(x',a,y))]\phi'(da)  + (\lambda- \lambda') c(w_i,y)}^2 }\\
&\leq\exp\crbk{\frac{\xi^2}{2} \sqbk{\sup_{y\in \W}\abs{\int_\A [r(x,a,y) + \alpha \bar u^*(f(x,a,y))]\phi(da)  - \int_\A\bar [r(x',a,y) + \alpha \bar u^*( f(x',a,y))]\phi'(da)}  + c_\vee\abs{\lambda- \lambda'} }^2 }\\
\end{aligned}$$
Consider \begin{equation}\label{eqn:unaware_lip_in_x_phi}\begin{aligned}
&\abs{\int_\A [r(x,a,y) + \alpha \bar u^*(f(x,a,y))]\phi(da)  - \int_\A [r(x',a,y) + \alpha \bar u^*( f(x',a,y))]\phi'(da)} \\
&\leq\abs{\int_\A [r(x,a,y) + \alpha \bar u^*(f(x,a,y))] -  [r(x',a,y) + \alpha \bar u^*(f(x',a,y))]\phi(da)}+\abs{ \int_\A [r(x',a,y) + \alpha \bar u^*( f(x',a,y))][\phi -\phi'](da)}\\
&\leq L|x-x'| +\min\set{\beta \TV{\phi-\phi'},L W_1(\phi,\phi')}
\end{aligned}\end{equation}
\begin{remark}
    As we will easily see from the reset of the proof, if $\cQ$ is set of measures with $\epsilon$ covers of cardinality $O(\epsilon^{-d})$ in either $W_1$ or total variation distance, for example $\cQ$ is a set of smoothly parameterized set of measures or $|\A|<\infty$, then the entropy integral will be finite, yielding a $n^{-1/2}$ convergence rate. However, in the following development, we will focus on the case where $|\A|<\infty$ to get concrete dependencies on the dimensions, diameters, and the size of the action space. 
\end{remark}
With $|\A|<\infty$, we conclude that $$E_\epsilon \exp[\xi(R_n(\bd{u},g_\theta) - R_n(\bd{u},g_{\theta'}))]\leq \exp\crbk{\frac{\xi^2}{2}\rho(\theta,\theta')}$$
where $$\rho(\theta,\theta') := L|x-x'| + \beta\TV{\phi-\phi'} + c_\vee|\lambda-\lambda'|$$which is a distance on $\X\times \cQ\times \Lambda$. This shows that the process $\set{R_n(\bd{u},g_\theta):\theta\in\Theta}$ is subgaussian w.r.t. $\rho$. 

\par Therefore, using Dudley’s entropy integral \citep[Chapter 5]{wainwright2019high_stats}, the empirical Rademacher complexity can be bounded $$\cR_n(\bd{w},\cG)\leq \frac{32}{\sqrt{n}}\int_0^{D_\vee}\sqrt{\log\cN(\epsilon;\Theta,\rho)} d\epsilon$$
w.p.1., where 
$$\overline{D}_\vee:=   L\diam(\X) + 2\beta + c_\vee\delta\inv \beta \geq \sup_{g,g'\in\cG}\rho(g,g').$$ 
In particular, as the r.h.s. is deterministic, we take expectation over $\bd{W} = \set{W_i:i=1,\ds ,n}\sim\mu_0^n$ to conclude that the population Rademacher complexity $$\cR_n(\cG)\leq \frac{32}{\sqrt{n}}\int_0^{D_\vee}\sqrt{\log\cN(\epsilon;\Theta,\rho)} d\epsilon$$satisfying the same bound. 
Moreover, the covering number \begin{equation}\label{eqn:unaware_covering_bd}
    \begin{aligned}
\cN(\epsilon;\Theta,\rho) &\leq \cN(\epsilon;\X,L|\cd|)\cd\cN(\epsilon;\cQ,\beta\TV{\cd})\cd\cN(\epsilon;\Lambda,c_\vee|\cd|)\\
&\leq \cN(\epsilon;\X,L|\cd|)\cd\cN(\epsilon;B_1^{|\A|},\beta\norm{\cd}_1)\cd\cN(\epsilon;\Lambda,c_\vee|\cd|)\\
&\leq \crbk{1+\frac{L\diam(\X)}{\epsilon}}^{d_\X}\crbk{1+\frac{2\beta}{\epsilon}}^{|\A|}\crbk{\frac{c_\vee\diam(\Lambda)}{\epsilon}+1}.
\end{aligned}
\end{equation}
where $B_1^{|\A|}$ is the $|\A|$ dimensional $\ell_1$-ball of radius 1, and its covering number bound follows from \citet[Example 5.8]{wainwright2019high_stats}. 
Therefore, the entropy integral $$\begin{aligned}
\int_0^{\overline D_\vee}\sqrt{\log\cN(\epsilon;\Theta,\rho)} d\epsilon &\leq \int_0^{\overline D_\vee}\sqrt{ (d_\X+|\A|+1)\log(\overline D_\vee/\epsilon)} d\epsilon\\
& = \frac{\sqrt{\pi}}{2} \overline D_\vee\sqrt{d_{\X}+|\A|+1}
\end{aligned}$$
\par Therefore, we conclude that by Proposition \ref{prop:bound_est_err_by_oprtr_gap} and the Rademacher complexity bound \eqref{eqn:bound_emp_process_by_Rad_cplx}, the estimation error $$\begin{aligned}
\norm{\hat u-\bar u^*} &\leq  \beta \norm{\overline{\bd{T} }(\bar u^*) - \overline \cT (\bar u^*) } \\
&\stackrel{(i)}{\leq}   2\beta \cR_n(\cG)+  \beta\sqrt{\frac{2g_\vee}{n}\log{\frac{1}{\eta}}}\\
&\leq \crbk{ 32\sqrt{\pi}\beta \overline D_\vee\sqrt{d_{\X}+|\A|+1} + \sqrt{2}\beta^{3/2}\sqrt{2\log{\frac{1}{\eta}}}}n^{-1/2}
\end{aligned}$$
where $(i)$ holds w.p. at least $1-\eta$. This bound implies the theorem as we note that $d_\X\geq 1$.

\subsection{Proof of Theorem \ref{thm:unaware_fk_sample_complexity}}

Again, we have that
$$ \begin{aligned}\norm{\overline{\bd{T}} (\bar u^*) (x)- \cT (\bar u^*)(x) } &\leq \sup_{x\in  \X,\phi\in\cQ}\abs{ \inf_{\psi\in \widehat \cP}\int_{\W} \int_{\A} [r(x,a,w) + \alpha \bar u^* (f(x,a,w))]\phi(da)\psi(dw)- \inf_{\psi\in \cP}\int_{\W} \int_{\A} [r(x,a,w) + \alpha \bar u^* (f(x,a,w))\phi(da)\psi(dw)}\\
&=\sup_{\theta\in\Theta}\abs{ \inf_{\psi\in \widehat \cP}\int_{\W} g_\theta(w)\psi(dw)- \inf_{\psi\in \cP}\int_{\W} g_\theta(w)\psi(dw)}
\end{aligned}$$
where $\Theta = \X\times\cQ$ and for $\theta = (x,\phi)$, $$g_\theta(w):= \int_{\A} [r(x,a,w) + \alpha \bar u^* (f(x,a,w))]\phi(da).$$
\par By \citet[Corollary 1]{duchi2021learning}, for $n\geq k\vee 3$, the r.h.s. satisfies w.p. at least $1-2\cN(\epsilon/3;\cG,\norm{\cd})e^{-t}$, 
$$ \sup_{g\in\cG}\abs{ \inf_{\psi\in \cP}\int_{\W}g(w)\psi(dw)- \inf_{\psi\in\widehat \cP}g(w)\psi(dw) }\leq 30\beta \epsilon$$
where $\cG:= \set{g_\theta:\theta\in\Theta}$ and $$\epsilon = n^{-\frac{1}{k'\vee 2}}c_k(\delta)^2\crbk{\frac{c_k(\delta)}{c_k(\delta)-1}\vee 2}\crbk{\frac{1}{k}+\sqrt{t+2\log n}}.$$
Therefore, choosing $t = \log(2\cN(\epsilon/3;\cG,\norm{\cd})/\eta)$, we have that w.p. at least $1-\eta$
$$\norm{\bd{\overline T}(\bar u^*) - \overline\cT(\bar u^*)} \leq 30\beta n^{-\frac{1}{k'\vee 2}}c_k(\delta)^2\crbk{\frac{c_k(\delta)}{c_k(\delta)-1}\vee 2}\crbk{\frac{1}{k}+\sqrt{\log(2\cN(\epsilon/3;\cG,\norm{\cd})+\log\frac{1}{\eta}+2\log n}}.$$
\par To bound the covering number, we recall \eqref{eqn:unaware_lip_in_x_phi}. 
Again, we can generalize to continuum settings. However, we focus on the finite action setting in this paper. In this case, we have that by \eqref{eqn:unaware_lip_in_x_phi}, $\theta\ra g_\theta(w)$ is uniformly 1-Lipschitz in the distance $$\rho((x,\phi),(x',\phi')) = L |x-x'| + \beta \TV{\phi-\phi'}. $$ This and the Lipschitz covering number bound \citep[Chapter 2.7.4]{van_der_vaart1996week_conv_and_emprical_meas} implies that
$$\begin{aligned}
\log\cN(\epsilon/3;\Theta,\rho)  
&\leq d_\X\log \crbk{1+\frac{3L\diam(\X)}{\epsilon\wedge 1}} + |\A|\log\crbk{1+\frac{6\beta}{\epsilon\wedge 1}}\\
&\leq d_\X\log \crbk{1+3L\diam(\X)} + |\A|\log\crbk{1+6\beta} + \frac{1}{2}\crbk{d_\X+|A|}\log n\\
\end{aligned}$$
where we handle the covering number of probability measures on $(\A,\TV{\cd})$ the same way as in \eqref{eqn:unaware_covering_bd} and the last inequality uses $\epsilon\geq n^{-1/2}$. 
\par Therefore, defining $\overline D:=d_\X\log \crbk{1+3L\diam(\X)} + |\A|\log\crbk{1+6\beta}$, we conclude that $$\begin{aligned}
\norm{\hat u - u^*}
&\leq 30\beta^2 n^{-\frac{1}{k'\vee 2}}c_k(\delta)^2\crbk{\frac{c_k(\delta)}{c_k(\delta)-1}\vee 2}\crbk{\frac{1}{k} + \sqrt{\overline D+\log\frac{1}{\eta} + 2(d_\X+ |\A|)\log n}} 
\end{aligned}$$
w.p. at least $1-\eta$. 
\section{Proofs for Section \ref{sec:LB}}
\subsection{Proof of Lemma \ref{lemma:LB_instance}}
Since $f$ and $r$ doesn't depend on $a$, we have that 
$$\bar u^*(x) = u^*(x) = x + \alpha\inf_{\psi\in\cP} \int_\W u^*(w)\psi(dw)$$
We guess that $ u(x) = x+c$ is the unique solution, and define $c' = \inf_{\psi\in\cP} \int_\W w\psi(dw)$. Then, we have
$$ x + \alpha\inf_{\psi\in\cP} \int_\W u(w)\psi(dw)= x+\alpha c + \alpha\inf_{\psi\in\cP} \int_\W w\psi(dw) = x+\alpha (c+c')$$
This shows that if we choose $c = \beta c'$, then $u$ is the unique solution.

\par Now we move on to show the lower bounds. For fixed $x\in \X$, we define a local version of the minimax risk
\[
\mathfrak{M}_n(\cU,\cK,x) = \inf_{K}\sup_{\mu\in\cU}E_{\mu^n}\abs{K(W_1,\ds,W_n)(x) - \cK(\mu)(x)}\leq \mathfrak{M}_n(\cU,\cK)
\]
which trivially lower bounds the uniform version. We will prove Theorem \ref{thm:Wd_lb} and \ref{thm:fk_lb} by showing the same lower bound for this local risk. 

\subsection{Proof of Theorem \ref{thm:Wd_lb}}
We apply Le Cam's technique to prove the lower bound. Recall the instance in Lemma \ref{lemma:LB_instance}. Fix $x\in [0,1]$, for any $\eta > 0$ and $\mu_0,\mu_1\in\cU$ s.t. whenever $|\cK(\mu_0)(x) - \cK(\mu_1)(x)|\geq 2\eta$, we have
\[
\mathfrak{M}_n(\cU,\cK,x)\geq \frac{\eta}{2}(1-\TV{\mu_1^n - \mu_0^n}). 
\]
We consider $\mu_0 = p_0\delta_{\set{1}} + (1-p_0)\delta_{\set{0}}$ with $p_0\leq \frac{1}{2}$. Then
$$ \begin{aligned}
\cK(\mu_0)(x)-x &= \beta\inf_{\psi\in\cP} \int_\W w\psi(dw) \\
&= \beta\sup_{\lambda\geq 0} - \lambda\delta +\int_{[0,1]}\inf_{y\in[0,1]}\crbk{y+\lambda(w-y)^2}\mu_0(dw) \\
&= \beta\sup_{\lambda\geq 0} - \lambda\delta + p_0\frac{4\lambda-1}{4\lambda}\1\set{\lambda \geq \frac{1}{2}} +  p_0\lambda\1\set{0\leq\lambda < \frac{1}{2}}\\
&= \beta\max\set{\sup_{\lambda\geq 1/2} p_0 - \lambda\delta  -\frac{p_0}{4\lambda}, \sup_{0\leq\lambda < 1/2} p_0\lambda - \lambda\delta}\\
&= \beta\max\set{ p_0 - \sqrt{p_0\delta},\frac{p_0-\delta}{2}}
\end{aligned}$$
It is not hard to see the max is always achieved by $p_0 - \sqrt{p_0\delta}$. 
\par So, if we construct the local alternative $\mu_1 = p_1\delta_{\set{1}} + (1-p_1)\delta_{\set{0}}$, then $\cK(\mu_1)(x)-x = \beta(p_1-\sqrt{p_1\delta})$. Therefore, choosing any $p_1 = p_0 + c$ with $\frac{1}{2}\geq c > 0$, we have $$\begin{aligned}
\abs{\cK(\mu_0)(x) - \cK(\mu_1)(x)} &= \beta\abs{c+\sqrt{\delta}\crbk{\sqrt{p_0} - {\sqrt{p_0+c}}  }}\\
&\geq \beta\inf_{\xi\in[p_0,p_0+c]}\abs{c - \sqrt{\delta}\frac{1}{2}\xi^{-1/2}c } \\
&\geq\beta\crbk{1-\frac{\sqrt{\delta}}{2\sqrt{p_0}} }c \\
&\geq \frac{\beta c}{2}
\end{aligned}$$
Hence, we can choose $p_1 = p_0 + 4 \beta\inv \eta$ when $\eta\le \frac{\beta}{8}$ to achieve separation $|\cK(\mu_0)(x)-\cK(\mu_1)(x)|\geq 2\eta$. 
\par By properties of TV-distance, KL, and $\chi^2$-divergence we have that $$\begin{aligned}
    \TV{\mu_1^n - \mu_0^n}&\leq \frac{n}{2}D_{\text{KL}}(\mu_1||\mu_0)\\
    &\leq\frac{n}{2}\chi^2(\mu_1||\mu_0)\\
    &\leq \frac{n}{2}\frac{(p_0-p_1)^2}{p_0(1-p_0)}\\
    &= \frac{8n\eta^2}{\beta^2 p_0(1-p_0)}
\end{aligned}$$
So, for all $n\geq 1$, we can choose $$\eta = \frac{\beta \sqrt{p_0(1-p_0)}}{4\sqrt{n}}\leq \frac{\beta}{8}.$$ 
With this $\eta$, we conclude that $\TV{\mu_1^n - \mu_0^n}\leq \frac{1}{2}$ and hence
$$\mathfrak{M}_n(\cU,\cK)\geq \mathfrak{M}_n(\cU,\cK,x)\geq \frac{\eta}{4} =  \frac{\beta\sqrt{p_0(1-p_0)}}{16}n^{-1/2}. $$
Since $p_0$ is arbitrary, we can choose the maximizer $p_0 = \frac{1}{2}$.  
\subsection{Proof of Theorem \ref{thm:fk_lb}}\label{a_sec:proof:thm:fk_lb}
We lower bound the uniform risk by
$$\mathfrak{M}_n(\cU,\cK)\geq \mathfrak{M}_n(\cU,\cK,0). $$
To achieve this, we would like to apply Duchi Theorem 3\cite{}. 
\par Notice that for two-point distribution $\mu$ with support $\set{0,1}$, $$\cK(\mu)(0) = \inf_{D_{f_k}(\mu'||\mu)\leq \delta}E_{\mu'}\beta Z = - \beta +\sup_{D_{f_k}(\mu'||\mu)\leq \delta}E_{\mu'}\beta(1- Z). $$
Here, $\beta(1-Z)$ has a two-point distribution on $\set{0,\beta}$ under $\mu$. Therefore, Theorem 3 of \citet{duchi2021learning} applies. Define $$p_k(\delta) = (1+k(k-1)\delta)^{-\frac{1}{k-1}},\quad \chi_k(\delta) = \frac{k(k-1)\delta}{2(1+k(k-1)\delta)}.$$ We obtain that with $n$ s.t. 
$$\sqrt{\frac{p_k(\delta)(1-p_k(\delta))}{8n}}\leq \frac{1-p_k(\delta)}{2}\wedge p_k(\delta), \quad \frac{1}{4n}\leq p_k(\delta)\wedge (1-(1-\chi_k(\delta))^{1-k'}p_k(\delta)),$$
then
$$\mathfrak{M}_n(\cU,\cK,0)\geq \beta \max \set{\frac{\sqrt{p_k(\delta)(1-p_k(\delta))}}{16\sqrt{2}k'p_{k}(\delta)}n^{-\frac{1}{2}}, \frac{\chi_k(\delta)^{\frac{1}{k}}c_k(\delta)}{8\cd 4^{k'}}n^{-\frac{1}{k'}}} . $$
This implies the statement of Theorem \ref{thm:fk_lb}. 

\section{Algorithm Design for the CAU Case}\label{a_sec:CAU_alg}
To parameterize randomized controller policies, we employ a generative model (Algorithm \ref{alg:cau}) by considering $\pi_\eta:\R^d \times \X\rightarrow \A$ where an action is generated by $A\sim \pi_\eta(N,x)$ using an independence source of randomness $N\sim N(0,I)$, a standard Gaussian vector independent of the state.

Under this definition, we overwrite the notation and consider
$$\widehat {\bd T} _ {\eta,\theta}(\lambda,x):=  \lambda - c_k(\delta)\sqbk{\int_{\W} (E_{N}[r(x,\pi_\eta(N,x)) + \alpha u_\theta(f(x,\pi_\eta(N,x),w))]-\lambda)_+^{k'}\hat\mu(dw) }^{1/k'}$$

By strong duality,  the Bellman operator under policy $\pi_\eta$ applied to $u_\theta$ is $$ \sup_{\lambda\in\R}\crbk{\lambda - c_k(\delta)\sqbk{\int_{ \W} \crbk{E_N [r(x,\pi_\eta(N,x),w) + \alpha u_\theta(f(x,\pi_\eta(N,x),w))]  - \lambda}^{k'}\hat\mu(dw)}^{1/k'} } = \widehat {\bd T}_{\eta,\theta}(\lambda^\ast,x)$$  where $\lambda^\ast = \lambda^\ast(\eta,\theta,x)$ is the optimal dual multiplier. We note that $\lambda^\ast$ doesn't depend on the realizations of $N$, and can be computed via bisection search.

\textbf{Bellman Error Minimization:} Analogous to the CAA case, we minimize the $L^2$ Bellman error:
$$\min_{\theta} \int_\X [u_\theta(x) - \widehat {\bd T}_{\eta,\theta}(\lambda^\ast,x)]^2 \nu(dx)$$
The gradient is evaluated using the envelope theorem $$ \nabla_\theta \bd T_{\eta,\theta}(x)=\nabla_\theta \widehat {\bd T}_{\eta,\theta}(\lambda^\ast(\eta,\theta,x),x)$$ and the expectation over $N$ can be approximated using the sample average over $m$ i.i.d. samples $N_1,\dots,N_m$.

Therefore, we update the $\theta$ using  first order methods; e.g. mini-batch stochastic gradient descent (SGD): $$\theta_ {t+1} = \theta_ t-\ell_ t G_ {n,m,t} $$ where $\ell_t$ is the learning rate and $G_{n,m,t}$ is a stochastic gradient. 

The gradient estimate  $G_{n,m,t}$ can be obtained as follows: we first sample  $\bd N=\set{N_1,\dots,N_m}$ and $X_i\sim \nu$ i.i.d. and compute $\lambda_{m,i}^\ast = \lambda^\ast(\eta,\theta,\bd N, X_i)$ that maximize
$$\lambda - c_k(\delta)\sqbk{\int_{ \W} \crbk{\frac{1}{m}\sum_{j=1}^mu_\theta(f(X_i,\pi_\eta(N_j,X_i),w))  - \lambda}^{k'}\hat\mu(dw)}^{1/k'} $$
using bisection search. Then, we set 
$$G_{n,m,t} = \frac{1}{n}\sum_{i=1}^n 2(u_\theta(X_i) - \widehat {\bd T} _ {\eta,\theta}(\lambda_{m,i}^\ast,X_i))(\nabla_\theta u_\theta (X_i)- \nabla_\theta\overline{\bd T} _ {\eta,\theta}(\lambda^\ast_{m,i},X_i) ).$$

\textbf{Policy Improvement:} The policy improvement step parallels the CAA setting:
\[
\eta_{t+1} = \eta_t + \ell'_t \frac{1}{n}\sum_{i=1}^n \nabla_\eta \widehat {\bd T}_ {\eta,\theta}(\lambda^\ast_{m,i},X_i),
\]
for some possibly different learning rate $\ell'_t$.

\begin{algorithm}[H]
\caption{Robust Actor–Critic Algorithm for CAU}\label{alg:cau}
\begin{algorithmic}[1]
\Require Step–size schedules $\{\ell_t\}_{t\ge0},\{\ell'_t\}_{t\ge0}$; state–batch size $n$; noise–sample size $m$;
         initial parameters $(\theta_0,\eta_0)$; sampling measure $\nu$ on $\X$
\vspace{0.3em}

\Statex \textbf{Repeat until convergence:}
\Statex \rule{\linewidth}{0.4pt}
\Statex \textbf{1. Bellman–error (critic) update}
\State Sample states $\{X_i\}_{i=1}^{n}\overset{\text{i.i.d.}}{\sim}\nu$
\State For each $i$, draw i.i.d.\ noise vectors $\{N_{ij}\}_{j=1}^{m}\sim\cN(0,I_d)$
\For{$i=1,\dots,n$ \textbf{in parallel}}
    \State $\displaystyle
        \lambda_i^{\!*}\;\leftarrow\;
        \arg\max_{\lambda\in\R}\;
        \widehat{\bd T}_{\eta_t,\theta_t}\!\bigl(\lambda,X_i;\{N_{ij}\}_{j=1}^{m}\bigr)$
        \Comment{bisection search}
\EndFor
\State $\displaystyle
    g_{\theta}\;\leftarrow\;\frac1n\sum_{i=1}^{n}
    2\Bigl(u_{\theta_t}(X_i)-\widehat{\bd T}_{\eta_t,\theta_t}(\lambda_i^{\!*},X_i)\Bigr)
    \Bigl(\nabla_{\!\theta}u_{\theta_t}(X_i)
          -\nabla_{\!\theta}\widehat{\bd T}_{\eta_t,\theta_t}(\lambda_i^{\!*},X_i)\Bigr)$
\State $\theta_{t+1}\;\leftarrow\;\theta_t-\ell_t\,g_\theta$
\vspace{0.5em}

\Statex \textbf{2. Policy–improvement (actor) update}
\State Resample (or reuse) $\{X_i\},\{N_{ij}\}$
\For{$i=1,\dots,n$ \textbf{in parallel}}
    \State Recompute $\displaystyle
      \lambda_i^{\!*}\leftarrow
      \arg\max_{\lambda}\widehat{\bd T}_{\eta_t,\theta_{t+1}}(\lambda,X_i;\{N_{ij}\})$
\EndFor
\State $\displaystyle
    g_{\eta}\;\leftarrow\;\frac1n\sum_{i=1}^{n}
    \nabla_{\!\eta}\widehat{\bd T}_{\eta_t,\theta_{t+1}}(\lambda_i^{\!*},X_i)$
\State $\eta_{t+1}\;\leftarrow\;\eta_t+\ell'_t\,g_\eta$
\State $t\;\leftarrow\;t+1$
\Statex \rule{\linewidth}{0.4pt}
\vspace{0.2em}
\Return final parameters $(\theta_{t},\eta_{t})$
\end{algorithmic}
\end{algorithm}
\section{Additional Experiment Results and Implementation Details}
\label{a_sec:result_exp}

\subsection{Inventory Control}

\subsubsection{Implementation Details}
\paragraph{Actor Network}
We use a 3-layer neural network with ReLU as the activation function, 8 neurons in each hidden layer. We treat $X_t$ as the CAA input, and $A_t$ as the output. With CAU, we have another standard normal random variable $N_1$ as the input.

\paragraph{Critic Network}
We use a 3-layer neural network with ReLU as the activation function, 32 neurons in each hidden layer. We treat $X_t$ as the input, and the value of the policy $u(X_t)$ as the output.

\paragraph{Training}
To achieve convergence of the actor-critic network, we choose a small batch size (1), a small learning rate ($3\times 10^{-4}$, Adam optimizer) and a large number of epochs ($5 \times 10^4$). For each of CAA and CAU cases, we train 5 sets of robust value and policy under $\delta \in \{1\times10^{-3}, 1\times10^{-2}, 1\times10^{-1}, 5\times10^{-1}, 1\}$.

\subsubsection{Discussions on Numerical Techniques}
\paragraph{Rescale the Data}
We rescale the demand of the good by multiplying $10^{-3}$ to the original data. This results in faster convergence of the robust value and policy because we could have a smaller sample space for $X_t$ than before. As a consequence, our reported test result should be multiplied by $10^3$.

\paragraph{Compact Sample Space}
Similar as the previous case, we need a compact support for $\nu(x)$ to ensure convergence. In this case, we observe that all the demand data lies in $[0, 10]$ after rescaling. Thus, we choose $\nu(x)$ to be uniform on $[0, 10]$.

\paragraph{Imposing i.i.d. Data Assumption}
In reality, the demand data may not be i.i.d. as we assume here. For example, the demand would surge during holidays and may have a seasonal pattern. To make the demand data as i.i.d. as possible, we only keep the data with features Open = 1, Promo=0, StateHoliday=0, SchoolHoliday=0, namely, the regular demand not during promotion or holiday.

\begin{figure}[t!]
    \centering
    \begin{subfigure}[t]{0.48\textwidth}
        \centering
        \includegraphics[height=2.5in]{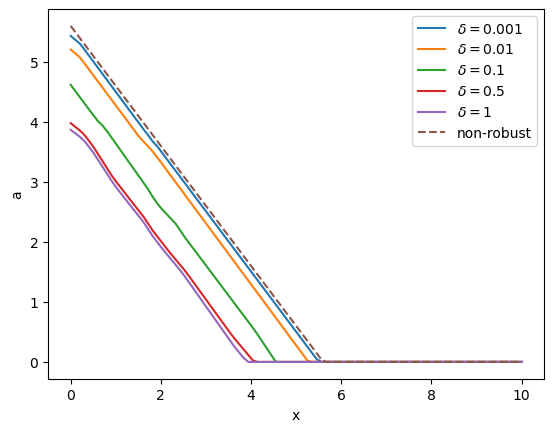}
        \caption{CAA policies for Store 1}
        \label{fig:inventory_caa_policy_1}
    \end{subfigure}
    \begin{subfigure}[t]{0.48\textwidth}
        \centering
        \includegraphics[height=2.5in]{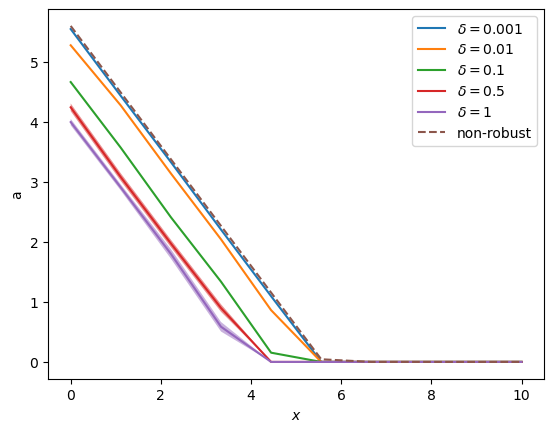}
        \caption{CAU policies for Store 1}
        \label{fig:inventory_cau_policy_1}
    \end{subfigure}
    \caption{Robust policies for one store in the inventory control problem.}
    \label{fig:inv_policy}
\end{figure}

\subsection{Portfolio Optimization}
\subsubsection{Implementation Details}
\paragraph{Actor Network}
We use a 3-layer neural network with ReLU as the activation function, 8 neurons in each hidden layer. For simplicity, we treat $\bd{1}^\top X_t$ as the CAA input, and $A_t^\top X_t$ as the output, rather than acquire a percentage $A_t$ first. With CAU, we have another two standard normal random variables $N_1$ and $N_2$ as the input.

\paragraph{Critic Network}
We use a 3-layer neural network with ReLU as the activation function, 32 neurons in each hidden layer. We treat $\bd{1}^\top X_t$ as the input, and $u(\bd{1}^\top X_t)$ as the output.

\paragraph{Training}
To achieve convergence of the actor-critic network, we choose a small batch size (1), a small learning rate ($3\times 10^{-5}$, Adam optimizer) and a large number of epochs ($10^5$). For each of CAA and CAU cases, we train 7 sets of robust value and policy under $\delta \in \{5\times10^{-4}, 1\times10^{-3}, 5\times10^{-3}, 1\times10^{-2}, 5\times10^{-2},1\times10^{-1},5\times10^{-1}\}$.

\subsubsection{Discussions on Numerical Techniques}
\paragraph{Compact Sample Space} Note that in our previous discussion, we should sample from $\nu(x)$ which is supported on $\R^2$. However, this would result in very slow convergence or no convergence at all. As a consequence, we choose a compact sample space $\{(x_1,x_2):x_1+x_2 \ge 0, x_1 \le 100, x_2 \le 100\}$. While this adjustment boosts convergence, it also induces bias in the robust value estimation. Empirically, we find the bias small and negligible. 

\paragraph{Enforcing Uniform Lipschitz Condition} In Section \ref{sec:UB} we assume the uniform Lipschitzness of the value function. However, in our setting, this is not the case in a small neighborhood around $\bd{1}^\top X_t =0$. Based on the form of the reward $r(X_t, A_t, W_t) = \frac{C_t^{1/2}}{1/2}$, we expect the robust value $u(\bd{1}^\top X_t)$ to behave like a square-root function. As a result, in order to enforce the uniform Lipschitz condition, we make the following change to the critic network: in the output layer, we multiply the output by $\sqrt{\bd{1}^\top X_t}$. This essentially results in a re-parametrized robust value function
$$ u_\theta(x) = v_\theta(x) \sqrt{\bd{1}^\top x}, $$
where $v_\theta(x)$ is the output of the critic network.

This makes the output of the critic network look like a constant function roughly. We find this technique particularly helpful with fitting the robust value and avoiding gradient explosion around $\bd{1}^\top x = 0$, where the Lipschitz condition is violated.

\paragraph{Action Clipping} In order to enforce the two constraints mentioned above: $A_t^\top X_t \ge 0$ and $A_t^\top X_t \le \bd{1}^\top X_t$, respectively, we clip the output of the actor network. If $A_t^\top X_t < 0$, then we make it to 0 by adding each output by $\frac{-A_t^\top X_t}{m}$; if $A_t^\top X_t > \bd{1}^\top X_t$, then we make it to $\bd{1}^\top X_t$ by decreasing each action by $\frac{A_t^\top X_t - \bd{1}^\top X_t}{m}$.


\end{document}